\documentclass[conference]{IEEEtran}
\usepackage{graphicx}
\usepackage{threeparttable}
\usepackage{epsfig}
\usepackage{epstopdf}
\usepackage{amssymb}
\usepackage{amsmath}
\usepackage{subfigure}
\usepackage[justification=centering]{caption}
\usepackage{amsfonts}
\usepackage{xcolor}

\usepackage{flushend}

\usepackage{booktabs}
\usepackage{url}
\usepackage{braket}
\usepackage{algpseudocode}
\usepackage{algorithm}
\usepackage{multirow}

\begin{document}
\title{Feature Importance and Explainability in Quantum Machine Learning\\
}

\author{\IEEEauthorblockN{Luke Power$^1$, Krishnendu Guha$^2$}
\textit{School of Computer Science and Information Technology, University College Cork, Ireland} \\
\textit{Email: 120371316@umail.ucc.ie$^1$ (corresponding author), kguha@ucc.ie$^2$}
 }
\maketitle

\IEEEtitleabstractindextext{%
\begin{abstract}
Many Machine Learning (ML) models are referred to as black-box models, providing no real insights into why a prediction is made. Feature importance and explainability are important for increasing transparency and trust in ML models, particularly in settings such as healthcare and finance. With quantum computing's unique capabilities, such as leveraging quantum mechanical phenomena like superposition, this can be combined with ML techniques to create the field of Quantum Machine Learning (QML), and such techniques may be applied to QML models. This article explores feature importance and explainability insights in QML compared to Classical ML models. Utilizing the widely recognized Iris dataset, classical ML algorithms— SVM and Random Forests, are compared against hybrid quantum counterparts, implemented via IBM’s Qiskit platform: the Variational Quantum Classifier (VQC) and Quantum Support Vector Classifier (QSVC). This article aims to provide an in-depth comparison of the insights generated in ML by employing permutation and leave-one-out feature importance methods, alongside ALE (Accumulated Local Effects) and SHAP (SHapley Additive exPlanations) explainers. 
\end{abstract}

}

\IEEEdisplaynontitleabstractindextext
\IEEEpeerreviewmaketitle

\section{Introduction}
This article aims to provide an understanding of quantum computing, and how it takes quantum mechanical phenomenon and integrates with classical machine learning to facilitate quantum machine learning (QML), alongwith explainability. \\ Given how many current ML models are considered 'black box' or 'opaque models', meaning that there is no real way to understand how or why an output is generated, there is a rise in the necessity to implement ways of explaining a models inner workings and rationalising its outputs. Currently, there are several methodologies to determine a model's most important features and explainability of individual predictions, and this article seeks to apply these methods to QML models and make an in-depth comparison of the results.

\subsection{Objectives}
\begin{figure}[!htb]
\centering
  \includegraphics[width=0.75\linewidth]{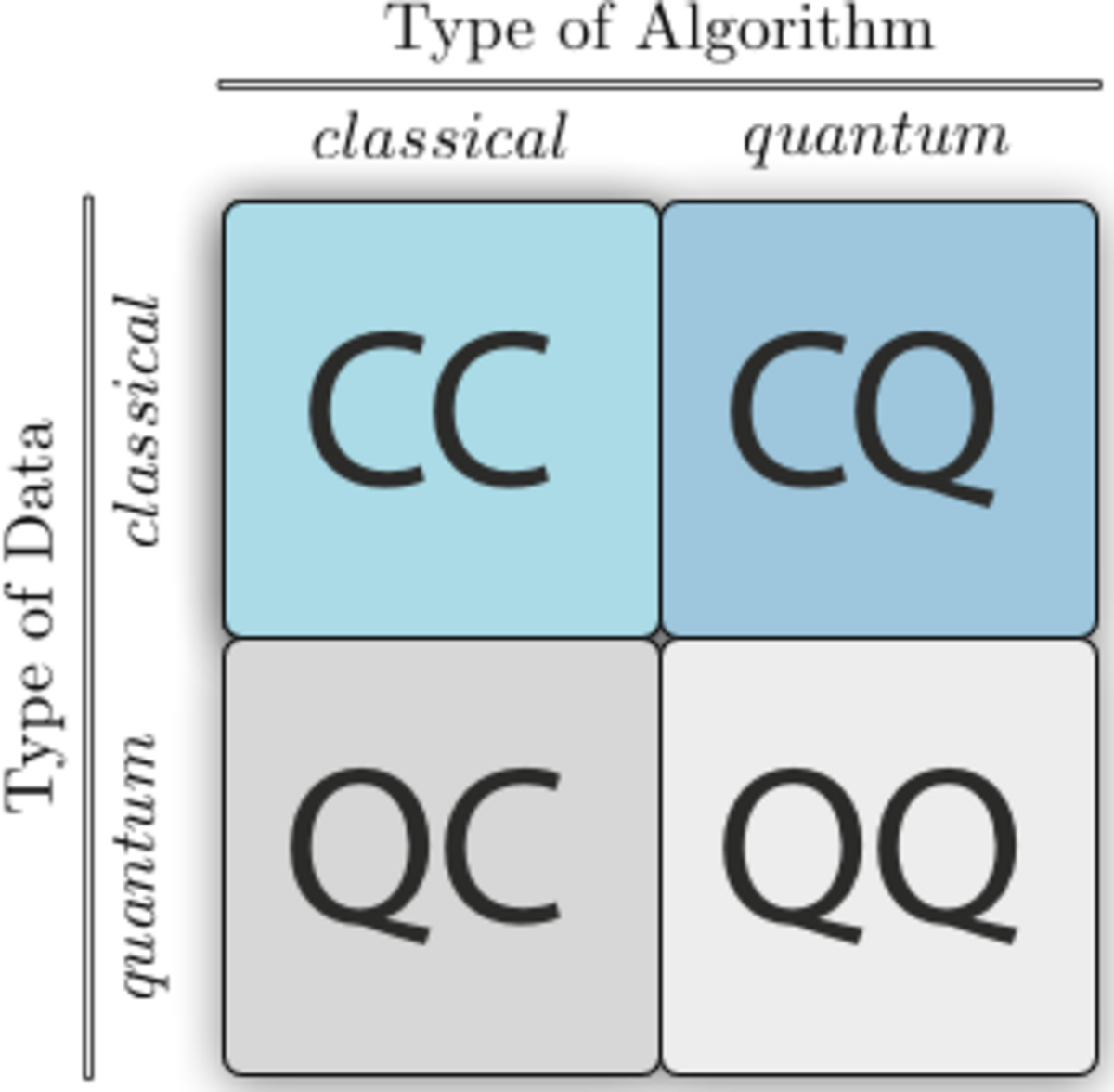}
  \caption[QML Approaches]{Classical Data on Quantum Algorithms. We are using Classical data with Quantum algorithms. Image adapted from \cite{schrodingerscatcartoon}.}
  \label{fig:Qml_approaches}
\end{figure}

The main objective of this article is to ascertain what new inferences could be made about a dataset via QML when compared to ML and attempt to explain how these differences come about. For this article, we are interested in using classical data, but quantum algorithms, in another way; the top-right quadrant of the image \ref{fig:Qml_approaches}. This was achieved by using a common ML method, building an equivalent QML model, and making inferences about the data by measuring feature importance - i.e. how important a feature is in making an accurate classification by using feature omission and permutations to quantify the differences, and also applying explainability methods to the models, which can provide greater insights into a singular prediction.\\ As of writing, this paper is the first to provide a comprehensive comparison of the results of applying machine learning explainability to QML on multi-class data. \\
This article will also delve into the theory behind quantum computing and the implementations of some basic quantum computing models to give clear demonstrations of the 'quantum advantage'.

\subsection{Work Overview}
The rest of the Introduction will provide a general overview of the article; the main objectives as well as some theoretical background on the main concepts of quantum computing and ML and a brief overview of the primary technology and tools that were used over the course of completing this article. \\
Chapter Two, Theoretical Background, will provide a more in-depth exploration of the machine learning models used, and explore the theory and mathematics behind implementing the QML models. This chapter will also demonstrate the implementation of quantum algorithms that provide some examples of quantum advantage in some computing tasks. The end of this chapter will also contain a literature review of the current research and papers in the field. \\
Chapter Three, Implementation goes through the actual implementation of the models on the Iris dataset and provides a comparison of the models' predictive performance, in the subsequent chapter, Results, the results of the implementations of the Feature Importance and Explainability methods are revealed, analysed and compared.\\
The Conclusions chapter will provide a personal reflection on the undertaking of this article, reviewing the experience in completing this article, how it was completed along with the future plans of the article.

\subsection{Overview of Methodology}
\begin{figure*}[!htb]
\centering
  \includegraphics[width=1\linewidth]{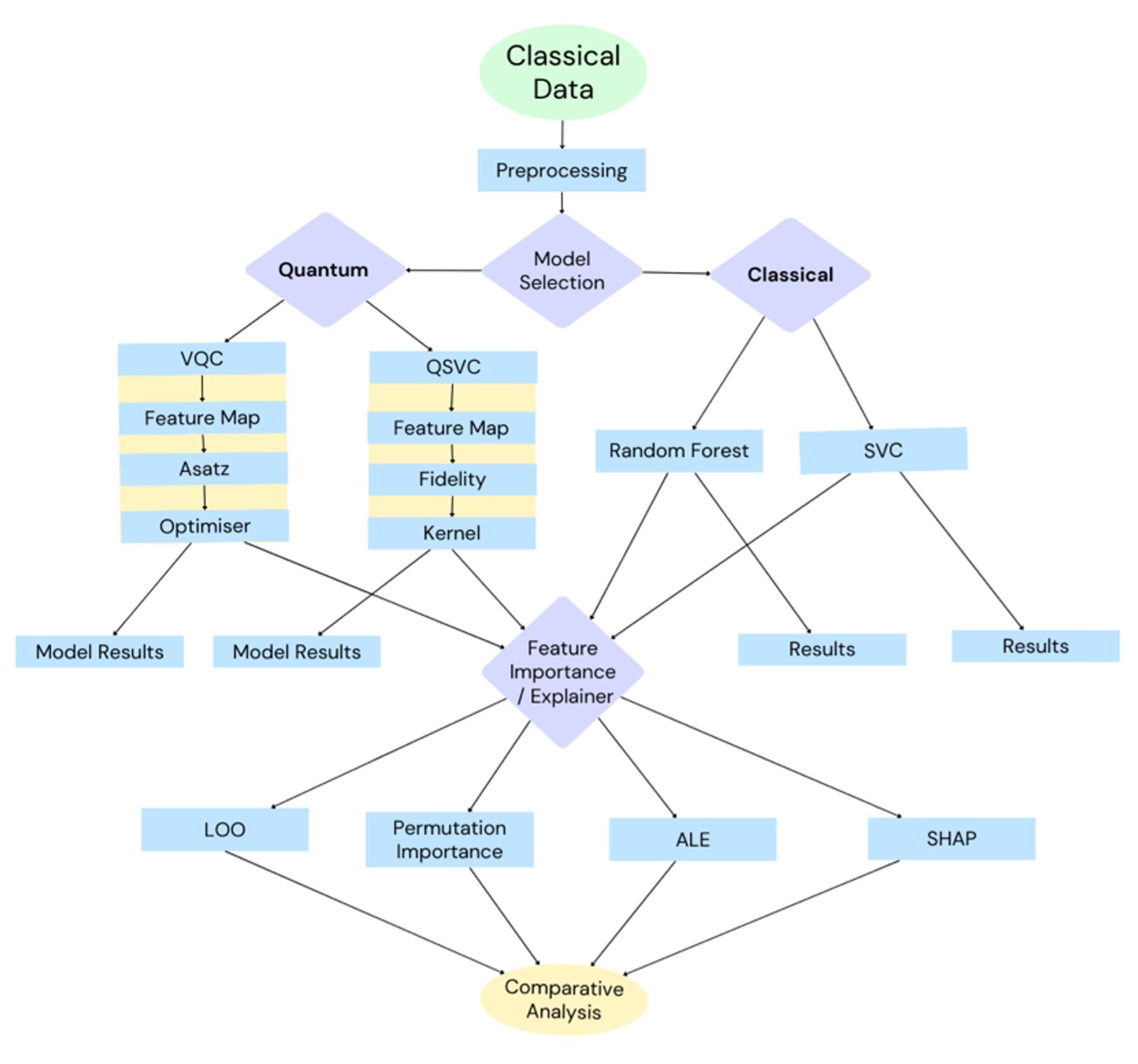}
  \caption[article Data Flow]{Data Flow of the article, centring around the application of Feature Importance and Explainability.}
  \label{fig:fyp_data_flow}
\end{figure*}

The approach to this article involved partaking in a course on quantum computing, familiarization with concepts of quantum theory, utilization of IBM's Qiskit platform which provides access to quantum hardware and building the necessary circuits and models. After several exploratory scripts and implementing algorithms which can provide proven demonstrations of the quantum advantage. After this, the initial QML models were built. The basic models were then expanded with different permutations of optimisers and ansatzes, to gain a level of accuracy comparable to the classical counterparts. After some initial model comparisons, we can then implement several methods of feature importance and explainability to both sets of models and compare the results. An overview of this process is displayed in the flow chart \ref{fig:fyp_data_flow}

\section{Background}

\subsection{Concepts and Technologies}
Machine Learning (ML) is a subsection of computer science and AI that utilizes statistical models to make inferences about the underlying patterns in a dataset and build algorithms that can make accurate predictions or classifications when presented with new, unseen data, without the need of explicit programming. \cite{Burkov2019}

Supervised learning, where models are trained on a labelled dataset, learning to predict outcomes for new data based on this training is the most widely used form of machine learning, and is the underlying method that was followed in the course of this article. Supervised learning involves feeding a model some labelled data which it will train its parameters on to minimise some cost function. 

 \subsection{Qubits and Superposition}
 QML is a branch of quantum computing that uses phenomena seen in quantum physics, such as superposition and entanglement which allows for multiple computations to be conducted simultaneously.\cite{Schuld_Sinayskiy_Petruccione_2014} The '\textbf{qubit}' is the quantum equivalent of the bit in classical computing. What is special about the qubit is that unlike bits, which can only ever have the states one or zero, a qubit can be in both states simultaneously. This phenomenon is known as \textbf{superposition}.


Schrodinger's Cat is a thought experiment to describe the idea of superposition, which is explained by thinking of a cat that is placed in a container containing a radioactive element that has a 50\% probability of releasing and killing the cat. It is said that until the box is opened and the state of the cat is observed (the state being either alive or dead), the cat is said to be in a state of superposition of both alive and dead at the same time\cite{Wineland_2013}. A more feline-friendly way to think about superposition is a coin that has been tossed in the air and is 'simultaneously heads and tails'. The ability to have data in superposition can allow for quantum algorithms to perform computations on data in multiple states simultaneously. This can lead to exponential speedups in computation and the ability to perform tasks on highly complex data. 

\subsection{Feature Importance \& Explainable ML}
Although ML models have provided great utility to many people and companies across a multitude of sectors, many of them are often regarded as 'black boxes'. The data is passed in and a result is returned, the actual inner workings of the model are not known to the user. This leads to an investigation into 'Explainable AI' XAI and 'Explainable ML', XML, which aims to shine a light on the inner workings of a model. Reasons for this could be in a denied loan application, and the ability to explain clearly why the model decided to deny the application, to a doctor or patient, explaining why a model predicted a certain medical diagnosis, or perhaps what settings are causing a machine to malfunction. Good explainability can also be utilised by data scientists to provide sanity checks or debugging on the model (what if someone's name was the reason behind the denied loan?). \cite{Belle_Papantonis_2021} \cite{Horel_2020}

\subsection{Methods for Assessing Feature Importance:}

\begin{itemize}
    \item \textbf{Leave-One-Out (LOO) Feature Importance}: This method assesses the impact on model performance when each feature is individually left out from the model. A significant decrease in performance upon leaving out a feature implies increased importance.
    
    \item \textbf{Permutation Feature Importance}: This method assesses feature importance by measuring the increase in the model's prediction error after permuting the feature's values, which breaks the relationship between the feature and the outcome.\cite{scikit}
    
    \item \textbf{SHapley Additive exPlanations (SHAP) Values}: Based on cooperative game theory, and Shapely values which in short, calculate how the winnings should be split among the team players, based on predictions on the outcomes had the players played independently or some permutations of them. This can be applied to ML through SHAP \cite{lundberg2017unified} values to explain the prediction of an instance by computing the contribution of each feature to the prediction. SHAP values offer both global and individual explanations of feature importance. \cite{SHAP}
    
    \item \textbf{Accumulated Local Effects (ALE)}: Used to interpret the influence of features in a model, especially in the context of prediction tasks. Unlike other feature importance metrics like in a Partial Dependency Plots, ALE values focus on the local effects of features, by splitting up the feature space into a number of 'windows' and calculating the average change in the model's predictions over the intervals of the feature's values. \cite{Molnar_2023}
\end{itemize}

    

The main software and package dependencies can be found in table \ref{tab:dependencies}.
\begin{table}
\centering
\caption{article Git Dependency}
\label{tab:dependencies}
\begin{tabular}{|l|l|l|}
\hline
\textbf{Package Name}      & \textbf{Version} & \textbf{License} \\ \hline
attrs                      & 21.4.0           & MIT              \\ \hline
brotlipy                   & 0.7.0            & MIT              \\ \hline
configparser               & 6.0.1            & MIT\\ \hline
cryptography               & 37.0.1           & Apache-2.0 and others\\ \hline
cython                     & 0.29.32          & Apache-2.0\\ \hline
dl                         & 0.1.0            & BSD-2-Clause\\ \hline
docutils                   & 0.18.1           & BSD-2-Clause\\ \hline
htmlparser                 & 0.0.2            & BSD-2-Clause\\ \hline
importlib-metadata         & 7.0.1            & Apache-2.0\\ \hline
ipython                    & 8.12.3           & BSD-2-Clause and BSD-3-Clause\\ \hline
ipywidgets                 & 7.6.5            & BSD-2-Clause and BSD-3-Clause \\ \hline
jinja2                     & 3.1.3            & BSD-2-Clause and BSD-3-Clause \\ \hline
jnius                      & 1.1.0            & MIT\\ \hline
keyring                    & 23.4.0           & MIT\\ \hline
lockfile                   & 0.12.2           & MIT\\ \hline
lxml                       & 4.9.1            & BSD-2-Clause and BSD-3-Clause \\ \hline
matplotlib                 & 3.5.2            & PSF-2.0\\ \hline
numpy                      & 1.23.5           & BSD-2-Clause\\ \hline
ordereddict                & 1.1              & MIT \\ \hline
pandas                     & 1.4.4            & BSD-2-Clause and BSD-3-Clause \\ \hline
pillow                     & 10.2.0           & HPND\\ \hline
pillow                     & 9.2.0            & HPND\\ \hline
protobuf                   & 4.25.3           & BSD-3-Clause\\ \hline
pyopenssl                  & 22.0.0           & Apache-2.0\\ \hline
pyopenssl                  & 24.0.0           & Apache-2.0\\ \hline
qiskit                     & 0.44.2           & \\ \hline
qiskit-aer                 & 0.12.2           & \\ \hline
qiskit-ibmq-provider       & 0.20.2           & Apache-2.0\\ \hline
qiskit-terra               & 0.25.2.1         &                   \\ \hline
railroad                   & 0.5.0            & MIT\\ \hline
scikit-learn               & 1.4.0            & BSD-2-Clause and BSD-3-Clause \\ \hline
seaborn                    & 0.11.2           & BSD-2-Clause and BSD-3-Clause  \\ \hline
sphinx                     & 5.0.2            & BSD-2-Clause                   \\ \hline
thread                     & 0.1.3            & BSD-2-Clause and BSD-3-Clause \\ \hline
toml                       & 0.10.2           & MIT\\ \hline
tornado                    & 6.1              & Apache-2.0 \\ \hline
xmlrpclib                  & 1.0.1            &                       \\ \hline
zipp                       & 3.8.0            & MIT\\ \hline
\end{tabular}
\end{table}


\section{Theoretical Background}
\subsection{Supervised Machine Learning}
Supervised ML consists of first gathering the labelled data, which the ML model will be trained and tested on. For classification tasks, this often consists of collecting data and filling in a number of feature vectors and a classifying label. Examples of this could be spam mail, where the features could consist of time, sender, and content keywords, among other possible predictors, and the classification label could be a simple 1 = spam, 0 = not spam. Another example could be a collection of photos, along with labels, classifying the image subjects, for example, differentiating cats and dogs. The feature vector in this regard would be the values of the image pixels' RGB and opacity values. \cite{Burkov2019}

There are a number of ML algorithms, but the one used for this article was the SVM. 

\subsubsection{Support Vector Machines}
Support Vector Machines are a powerful ML tool that can be used for both regression and classification but is more often used in classification tasks. An SVM works by finding an optimal hyperplane that separates the data points with the best probability of correctly separating classes where there is overlap by maximising the margins of the hyperplane.


The separating hyperplane\(f(x)\), defined by:  \[f(x) = \beta_0 + \beta_1 X_{1i} + \beta_2 X_{2i} + \ldots + \beta_p X_{pi}\] for a given point \((x_i,y_i)\) is so that: \[f(x_i) = \begin{cases}
            > 0, & \text{if \(y_i = 1\)}\\
            < 0, & \text{if \(y_i = 0\)}
\end{cases}\]
And we can use this as a classification rule for the data; classifying \(x^*\) with respect to \((sign)f(x^*)\)

The objective of the SVM is to calculate the optimal hyperplane to separate the data defined by \[ M = argmax_{\{\beta,\epsilon,m\}}\\\{y_i(\beta_0 + \beta_1 x_{1i} + \ldots + \beta_p x_{pi}) \ge m(1-\epsilon_i\}
\]

Subject to: \[
\sum \beta_j^2 = 1, \quad \epsilon_i \ge 0, \quad \sum_{i = 1}^{N} \epsilon_i \le c
\]

For some tuning parameter \(c > 0\), also known as a 'budget' which is the allowed rate for margin violation, and \(\epsilon_i\) or 'slack variable', where:
\begin{flalign}
\epsilon_i > 0 & \quad \text{point is classified on the wrong side of the  margin}, \\
\epsilon_i > 1 & \quad \text{point is classified on wrong side of the hyperplane}.
\end{flalign}

\begin{figure}[!htb]
  \centering
  \includegraphics[width=0.8\linewidth]{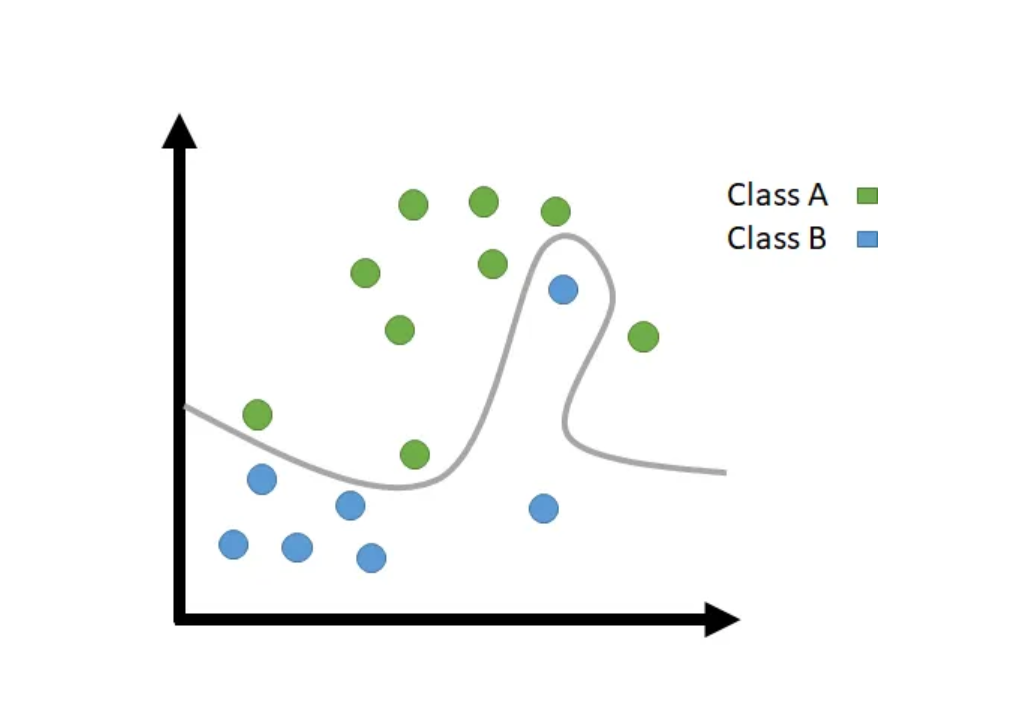}
  \caption[Support Vector Machine with Kernel trick]{Non-Linear Support Vector Machine. Image adapted from \cite{nonlinearsvm}.}
  \label{fig:nonlinearsvm}
\end{figure}

In real life, data is often non-linearly separable, and the formula for the hyperplane needs to be expanded to account for non-linearity by incorporating functions known as kernels on the predictors, as seen in figure \ref{fig:nonlinearsvm}. 

This introduction of kernel function is known as the 'kernel trick', which updates the formulas for the hyperplane as follows: 

\[ M = argmax_{\{\beta,\epsilon,m\}}\\\{y_i(\beta_0 + \sum_{j = 1}^P (\beta_j x_{i_i} + \alpha_j x_{j_i}^2) \ge m(1-\epsilon_i\}
\]

For all \(i\), subject to: \[
\sum\epsilon_i \le c,
\quad \epsilon_i \ge 0, 
\quad c > 0, 
\quad \sum \beta_j^2 + \sum_j \alpha_j^2 = 1
\]

We can then establish that an SVM uses kernel functions \[
f(x) = \sum_{h=1}^H \alpha_h K_h(x,x') + \alpha_0 \]
\[
= \alpha_0 + \sum_{i \in \delta} \alpha_i k(x,x_i)
\]
Where \(\delta\) is the set of Support Vectors. \cite{Wolsztynski_SVM}

\subsubsection{Random Forest}

A random forest is an ensemble model that samples from a number of different decision trees and aggregates the output of the trees. This is due to the fact that while growing a tree, the optimal cutoff is on a split-by-split basis. This greedy approach can result in a sub-optimal tree. This strategy aims to reduce this effect, reduce variability and to preserve a low bias. In order to aggregate the results from the trees requires resampling. For example: using a bootstrap resampling scheme: 

\[
\hat{f}_{B}^{*}(x) = \frac{1}{B} \sum_{b=1}^{B} \hat{f}_{b}^{*}(x)
\]

Where \( B \) represents the number of bootstrap samples and \( \hat{f}_{b}^{*}(x) \) represents an estimate from each bootstrap sample. The notation \( \hat{f}_{B}^{*}(x) \) denotes the averaged estimate across all bootstrapped samples.\cite{Wolsztynski_RF}


\subsection{Quantum Machine Learning}
QML can enhance machine learning by leveraging quantum mechanics to identify patterns in data that classical computing struggles with. Research suggests that quantum computers might excel in tasks like pattern recognition and optimization due to their ability to process complex, counter-intuitive patterns. Although some literature indicates that quantum-enhanced machine learning could unlock new paradigms in data analysis and problem-solving, the biggest advantage of quantum computing is the speed-up that can be expected, particularly through quantum computers' ability to handle large matrix operations. \cite{Biamonte2017}\cite{Huang2022} \\

\begin{figure}[!htb]
  \centering
  \includegraphics[width=01\linewidth]{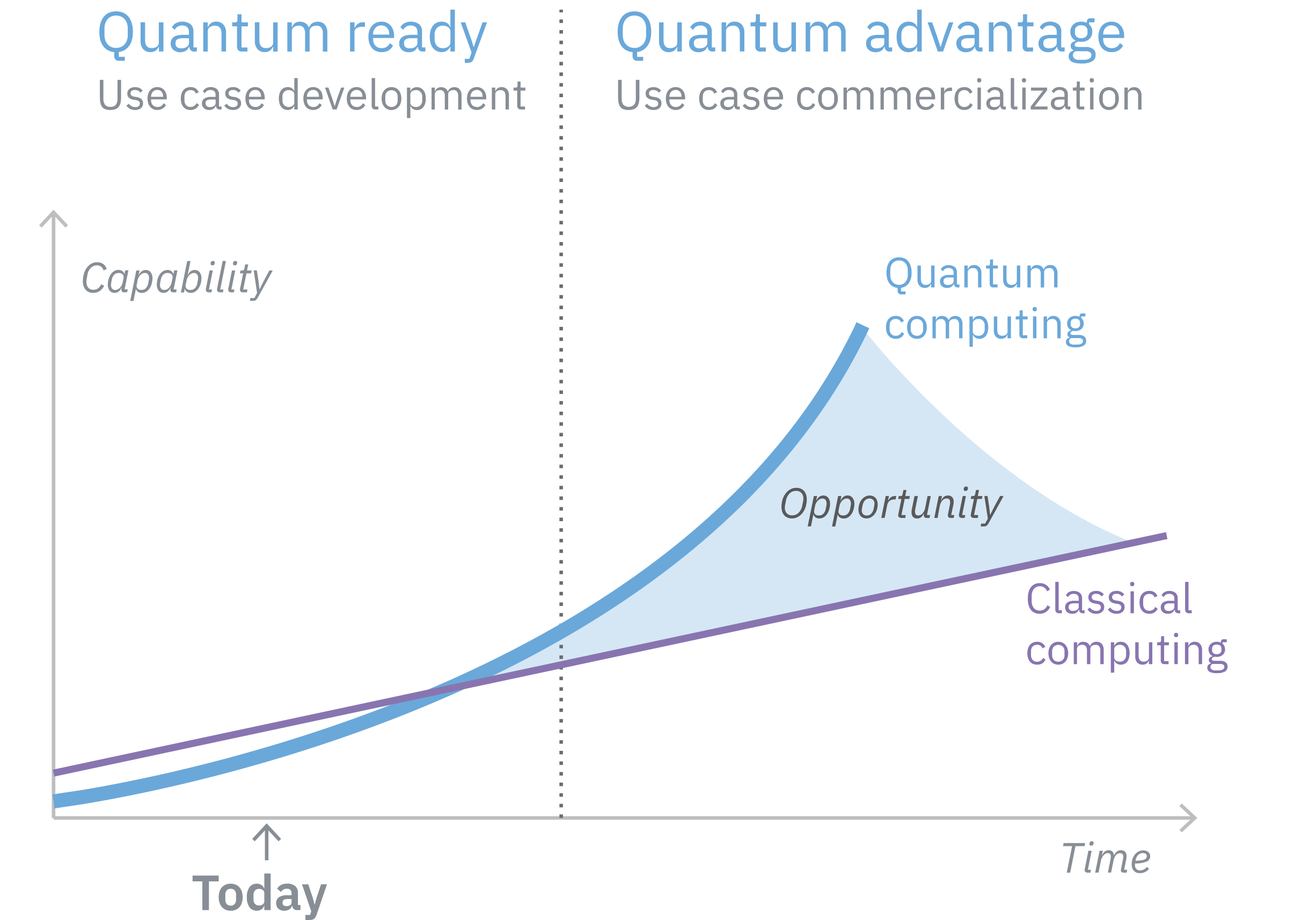}
  \caption[Quantum Advantage]{Quantum Advantage. Image adapted from \cite{quantum_advantage_ibm}.}
  \label{fig:quantumadvantage}
\end{figure}
The core advantage of quantum computing lies in its ability to process information in ways that classical systems cannot match, offering exponential speed-ups for certain types of problems. For example, quantum algorithms have been demonstrated to outperform classical counterparts in oracle-based problems, showing that a significant quantum advantage emerges even in existing noisy systems.\cite{Diego2017} Furthermore, quantum technology can revolutionize how we learn from experiments, as shown in a study where quantum machines learned from exponentially fewer experiments than required by conventional means, leading to dramatic reductions in the number of necessary experiments.\cite{Huang2022} Quantum machine learning also benefits from the ability to process atypical patterns produced by quantum systems, offering potential speed-ups in learning tasks.\cite{Biamonte2017} Additionally, the use of quantum-enhanced feature spaces in machine learning can provide advantages in solving classification problems where classical feature spaces become computationally prohibitive \cite{Havlicek2019}.

\subsubsection{The Qubit \& Superposition}
As mentioned in the introduction, quantum computing utilises qubits which differ from classical bits by being able to exist as both zero or one simultaneously in a state called superposition. This means that \( n \text{ bits can represent } n^2 \text{ possible states, but qubits can be in } 2^n \) states simultaneously. Superposition allows operations to be applied on all the possible states simultaneously, and combined with the likes of interference effects which can cancel out 'wrong' results can provide results that cannot be achieved with classical computing. \cite{QiskitSummerSchool}. Quantum computing also relies on linear algebra to mathematically represent states as vectors. For example, the classical bit's states of zero or one  would be represented as the vectors (using Dirac notation):
\[
 \ket{1} = \begin{bmatrix}
 0 \\ 1 \end{bmatrix} \text{and } \ket{0} = \begin{bmatrix}
     1 \\ 0 \end{bmatrix}
\]

In quantum computing, these certain states would be referred to as the two 'pure states'.
A qubit, on the other hand, would be in a combination of the two states simultaneously, in a phenomenon known as superposition taking on coefficients \(\alpha\) and \(\beta\) which represent the probabilities of being in their respective states. A qubit in a state \(\ket{\psi}\) would be written as;
\[
\ket{\psi} = \alpha\ket{0} + \beta\ket{1} , \quad \alpha,\beta \in C
\]

Where \(C\) represents complex numbers; numbers that have an imaginary component \(i \text{ where } i = \sqrt{-1}\). It is not possible to measure \(\alpha \text{ or } \beta \), as the observation or measurement of a qubit will cause it to collapse into one of its pure states. However since \(\alpha \text{ and } \beta\) are still probabilities their total magnitude must be equal to one. \[
|\alpha|^2 + |\beta|^2 = 1 \] \cite{Pattanayak_2021}\\

A superpositioned state can be achieved by passing qubits through a unitary operation known as a Hadamard gate. 

\begin{itemize}
    \item Applied to \(|0\rangle\), the Hadamard gate produces the state \(|+\rangle = \frac{1}{\sqrt{2}}(|0\rangle + |1\rangle)\), an equal superposition state where the probabilities of measuring the qubit in state \(|0\rangle\) or \(|1\rangle\) are both 50\%.\\
    \item Applied to \(|1\rangle\), it produces the state \(|-\rangle = \frac{1}{\sqrt{2}}(|0\rangle - |1\rangle)\), another equal superposition state, but with a phase difference between the \(|0\rangle\) and \(|1\rangle\) components. 
\end{itemize}
\cite{QiskitSummerSchool}\\

\begin{figure}[h!]
  \centering
  \includegraphics[width=0.5\linewidth]{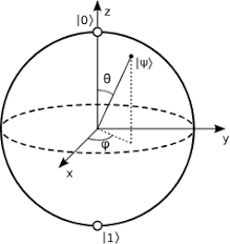}
  \caption[Bloch Sphere]{3D representation of a qubits’ possible states.}
  \label{fig:bloch_sphere}
\end{figure}

To better understand the state of a qubit we use what is called a 'Bloch Sphere', depicted in figure \ref{fig:bloch_sphere}. A qubit's state can be represented geometrically on a Bloch sphere described with the formula: 
\[ |\psi\rangle = \cos\left(\frac{\theta}{2}\right)|0\rangle + e^{i\phi}\sin\left(\frac{\theta}{2}\right)|1\rangle \]

In this representation, \( |0\rangle \) and \( |1\rangle \) are the standard basis states of the qubit. The angles \( \theta \) and \( \phi \) are real numbers, corresponding to the point on the Bloch sphere, where \( \theta \) ranges from 0 to \( \pi \) and \( \phi \) ranges from 0 to \( 2\pi \). The state \( |\psi\rangle \) is thus a point on the surface of the sphere, where the north pole corresponds to \( |0\rangle \) and the south pole to \( |1\rangle \). \cite{glendinning2005bloch}

\subsubsection{Variational Quantum Classifier}
A VQC represents an integration of quantum computing principles into machine learning. The process of implementing a VQC involves encoding classical data to a quantum feature space via a feature map, and then using this new quantum data in the quantum circuit. This allows for classical data to be used in a quantum circuit. A VQC uses variational quantum algorithms which follow hybrid quantum-classical schemes, involving parameterised circuit and gates with parameters optimised through a classically-based optimisation loop to minimise some cost function. These steps are repeated in a quantum-classical hybrid loop that eventually terminates when the classical optimization has found optimal parameters. The typical cost function is a measurement between the actual outputs and the desired outputs for training data. \cite{QiskitSummerSchool} \cite{Qiskit2021}

\begin{figure*}[h!]
  \centering
  \includegraphics[width=0.8\linewidth]{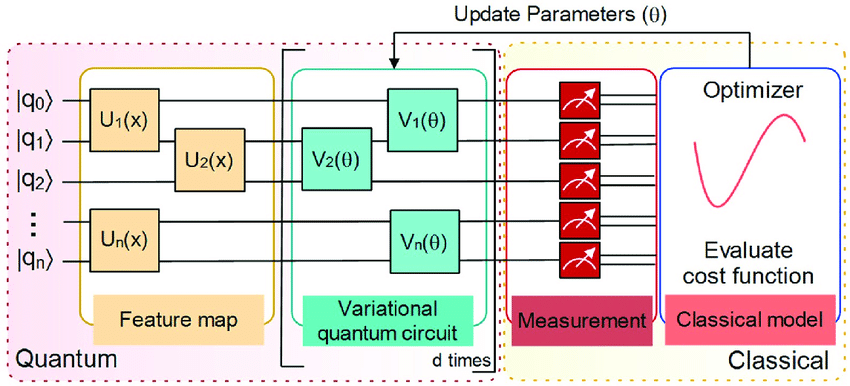}
  \caption[VQC Schematic]{Schematic representation of a variational quantum circuit (VQC). Image from \cite{vqc_schematic}}.
  \label{fig:VQC_Schematic}
\end{figure*}

\subsubsection{Quantum Support Vector Machines}
Quantum Support Vector Machines are the quantum extension to SVMs discussed earlier by offering a natural and efficient framework for linear algebra operations, including those required for SVMs. QSVMs exploit quantum algorithms for linear systems of equations and quantum kernel estimation to calculate the inner products in a high-dimensional feature space more efficiently. This approach leverages the concept of quantum feature maps and kernels to implicitly perform these calculations in a Hilbert space represented by quantum states. What this suggests is that QSVMs' quantum kernels are expected to do better than classical if they are hard to estimate classically.\cite{QiskitSummerSchool}

The quantum kernel is created by mapping a classical feature vector $\vec{x}$ to a Hilbert space using a quantum feature map $\phi(\vec{x})$. 
\[
K_{ij} = |\langle \phi(\vec{x}_i) | \phi(\vec{x}_j) \rangle|^2
\]

where $K_{ij}$ is the kernel matrix, $\vec{x}_i, \vec{x}_j$ are $n$ dimensional inputs $\phi(\vec{x})$ is the quantum feature map $|\langle a | b \rangle|^2$ denotes the overlap of two quantum states $a$ and $b$.\cite{kernel_QML}

\subsection{Encoding Classical Data}
One of the first steps in building the QML model is encoding (or 'embedding) the classical data into a quantum state. This step allows us to utilise phenomena like superposition and entanglement on the embedded classical data. \cite{QiskitSummerSchool} there are a number of ways of implementing this;

\begin{itemize}
    \item \textbf{Basis Encoding:} 
    Basis (or computational) encoding maps each bit of classical binary data to the corresponding quantum state. A classical bit value of 0 or 1 is represented by the quantum state $\ket{0}$ or $\ket{1}$ respectively. \cite{rath2023quantum}
    
    \item \textbf{Amplitude Encoding}
    Where data vector \(\Vec{x}\) in $\mathbb{R}^N$ is normalised and then encoded into the amplitudes of a quantum state $\ket{\psi}$, represented as: 
    \[
    \ket{\psi} = \sum_{i=0}^{N-1} x_i \ket{i},
    \]  where $\ket{i}$ denotes the computational basis states. This method allows for encoding $N$-dimensional data into $\log_2(N)$ qubits. \cite{Schuld}
    
    \item \textbf{Angle Encoding:} 
    Utilizes the values of classical data to define the angles in quantum rotations about the Bloch Sphere \cite{QiskitSummerSchool}. For instance, a real number $\theta$ is encoded as $\ket{\psi(\theta)} = R(\theta)\ket{0} + R(\theta)\ket{1}$, letting R represent the rotation procedure. For example a rotation about the Z axis would be represented as: The rotation matrix around the Z-axis is given by:
\[
R_z(\theta) = e^{-i\theta Z/2} = 
\begin{bmatrix}
e^{-i\theta/2} & 0 \\
0 & e^{i\theta/2}
\end{bmatrix}
\]
\cite{rath2023quantum}
\end{itemize}

\subsubsection{ZZ-Feature Map}
The  ZZFeatureMap, is a second-order Pauli-Z evolution circuit,\cite{QiskitDocs}
The Pauli Expansion circuit is a data encoding circuit that transforms input data $\vec{x} \in \mathbb{R}^n$, where $n$ is the number of feature dimensions, as

\[
U_{\Phi}(\vec{x}) = \exp \left( i \sum_{S \in I} \phi_{S}(\vec{x}) \prod_{i \in S} P_i \right).
\]

Here, $S$ is a set of qubit indices that describes the connections in the feature map, $I$ is a set containing all these index sets, and $P_i \in \{ I, X, Y, Z \}$. Per default the data-mapping $\phi_S$ is

\[
\phi_{S}(\vec{x}) = 
\begin{cases} 
x_i & \text{if } S = \{i\} \\
\prod_{i \in S} (\pi - x_i) & \text{if } |S| > 1
\end{cases}
\]
\cite{Havlicek2019}

The construction of the ZZFeatureMap involves applying a series of controlled rotation gates that correlate qubit states in a manner that reflects the underlying structure of the classical data.

\begin{figure}[h!]
  \centering
  \includegraphics[width=0.8\linewidth]{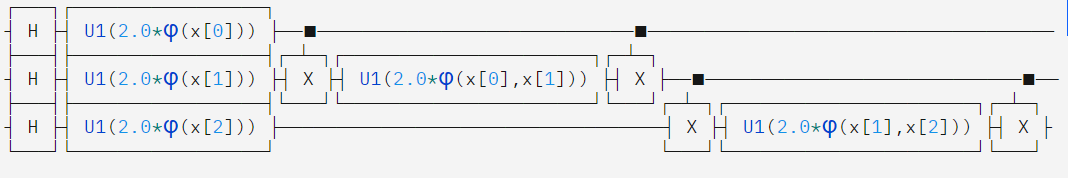} \caption[ZZFeatureMap]{3 qubits and 1 repetition and linear entanglement circuit.}
  \label{fig:zzfeaturemap}
\end{figure}

\ref{fig:zzfeaturemap} represents the circuit of the ZZFeatureMap where where \(\phi\) is a classical non-linear function, which defaults to \(\phi(x) = x \text{ if } \phi(x,y) = (\pi - x)(\pi - y)\).\cite{QiskitDocs}

\subsection{Feature Importance \& Explainability}
Feature importance plays a pivotal role in the domain of ML, serving as a bridge to understanding the inner workings of black box predictive models, which are often hidden from the user. This concept highlights the relative significance of each input feature in contributing to the model's ability to make accurate predictions. The importance of being able to understand a model spans various aspects of development and application, emphasizing the necessity for interpretability, transparency, and trust in machine learning outcomes.\\
In many critical sectors such as healthcare, finance, and more, decisions made by machine learning models can have profound implications. \cite{Horel_2020} For example in healthcare, a model might be used to predict patient outcomes based on various clinical parameters, and this outcome needs to be fully understood by not only the doctor but also to be explained to the patient. Understanding which features most significantly influence the predictions can provide insights into diseases, patient management, and treatment planning. Similarly, in finance, models predicting credit risk can benefit from clear insights into which factors most influence a person's likelihood of qualifying for a loan, aiding in fair and transparent decision-making. \\ The emphasis on feature importance thus directly ties into the goal of XAI. XAI seeks to peel back the layers of complex ML models to offer human-understandable explanations for their predictions. \\
With the increasing deployment of ML models in sensitive and impactful domains, regulatory bodies are mandating greater transparency and explainability in automated decision-making systems. For instance, the European Union's General Data Protection Regulation (GDPR) introduces the right to explanation, where individuals can ask for explanations of automated decisions that affect them. Feature importance metrics can help in fulfilling such regulatory requirements by providing a basis for explaining decisions made by ML models. \cite{das2020opportunities}\\

From a technical perspective, understanding feature importance is integral to refining and optimizing ML models. By identifying and focusing on the most 
features, data scientists can streamline models, reducing their complexity without sacrificing performance. This process of feature selection helps in mitigating overfitting, enhancing the model's ability to generalize to unseen data. Moreover, it can lead to more efficient models by reducing the dimensionality of the data, thereby lowering computational costs and improving runtime efficiency. \\ 

For machine learning solutions to be widely adopted, especially in high-stakes domains, it is crucial for end-users and stakeholders to trust the decisions made by these systems. Feature importance metrics facilitate this trust by offering transparency in the decision-making process. When users understand why a model makes a certain prediction, their confidence in the system's reliability and fairness should increase. This trust is paramount, especially in critical applications, where the cost of errors is high. \\

Feature importance and explainability are cornerstones of ethical and effective machine learning practice. It not only fosters trust, transparency, and ethical decision-making but also contributes to the technical robustness and efficiency of predictive models. As machine learning continues to permeate various sectors of society, the role of feature importance in ensuring responsible AI cannot be overstated. This understanding is achieved through various methods, including:

\begin{itemize}
    \item \textbf{Model Interpretability and Transparency}: By understanding which features significantly influence the model's output, stakeholders can gain insight into the model's decision-making process, leading to greater trust.
    
    \item \textbf{Improved Model Performance}: Identifying and focusing on important features can lead to more efficient models by reducing dimensionality, which can, in turn, lower computational costs and mitigate the risk of overfitting.
    
    \item \textbf{Feature Engineering}: Knowing which features are important can guide the feature engineering process, where new features are created and redundant or irrelevant features are removed.
    
    \item \textbf{Domain Insight}: Feature importance can reveal unexpected relationships between features and the target variable, providing new insights into the domain area being studied.
    
    \item \textbf{Regulatory Compliance}: In many regulated industries, the ability to explain decisions made by machine learning models is a legal requirement.
    \cite{Belle_Papantonis_2021} \cite{Horel_2020} .
\end{itemize}

\subsubsection{Leave-One-Out (LOO) Feature Importance}
Leave-one-out feature importance is an intuitive approach where the importance of a feature is determined by the impact on model performance when that feature is left out of the training process. In this method, the model is trained multiple times, each time leaving out a different feature. The change in model performance (such as accuracy or F1 score) indicates the importance of the omitted feature. If the performance drops significantly without a particular feature, it suggests that the feature is important for the model to make accurate predictions. However, this is a very time-consuming method as it requires retraining a model for each feature to be tested.

\subsubsection{Permutation Importance}
Permutation importance ranks the importance of a feature by shuffling the values of the column in question and recalculating the accuracy of the model. The permutation importance algorithm \ref{alg:perm_importance} shows how it is implemented. Permutation Importance breaks the relationship between the feature and the true outcome, thus the change in model error after permutation is indicative of the feature's predictive power. The main advantage of permutation importance is that it can be applied to any model and is less likely to be biased towards features with a high number of categories.\cite{scikit}

\begin{figure}[h!]
  \centering
  \includegraphics[width=0.8\linewidth]{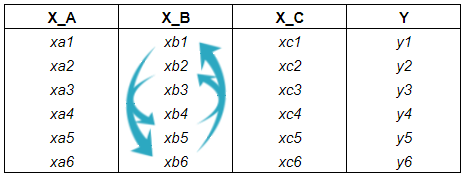}
  \caption[Permutation Importance]{How permutation importance works. Image from \cite{perm_importance}}
  \label{fig:perm_importance}
\end{figure}

\begin{algorithm}
\caption{Permutation Importance}\label{alg:perm_importance}
\begin{algorithmic}[1]
\Require fitted predictive model \(m\), tabular dataset \(D\) (training or validation)
\State Compute the reference score \(s\) of the model \(m\) on data \(D\)
\For{each feature \(j\) in columns of \(D\)}
\For{each repetition \(k = 1, \dots, K\)}
    \State Randomly shuffle column \(j\) of dataset \(D\) to generate a corrupted version of the data named \(\widetilde{D}_{k,j}\)
    \State Compute the score \(s_{k,j}\) of model \(m\) on corrupted data \(\widetilde{D}_{k,j}\)
\EndFor
    \State Compute importance \(i_j\) for feature \(f_j\) defined as: 
     \(i_j = s - \frac{1}{K} \sum_{k=1}^{K} s_{k,j}\)
\EndFor
\end{algorithmic}
\end{algorithm}

\subsubsection{SHAP Values} 
The SHAP method uses game theory to explain a machine learning model's output. Through the use of the traditional Shapley values from game theory and their related extensions, it links local explanations for optimal credit allocation.SHAP values give an overall measure of feature importance but also show the direction and magnitude of a feature's effect on individual predictions \cite{lundberg2017unified}. This method assigns each feature an importance value for a particular prediction by comparing what a model predicts with and without the feature. The calculation considers all possible combinations of features, which ensures fair attribution since it accounts for interaction effects between features. The formula can take the form: \[
\mathbb{E}[f(X) \,|\,do(X_S = x_S)]
\]

This tells us how the model would behave if the inputs were changed, and also because it is much easier to compute as computing SHAP values are generally NP-hard \cite{SHAP}.

\subsubsection{Accumulated Local Effects}
 Accumulated Local Effects plots calculate the local impact of a feature by evaluating the change in predictions over finely segmented windows of the feature space. This approach allows for a more accurate representation of the feature's effect, taking into account the natural interactions and dependencies among features. ALE plots achieve this by focusing on the differences in model predictions within these intervals, thus isolating the specific contribution of each feature to the prediction outcome. ALE plots average the changes in the predictions and accumulate them over the grid. This algorithm was introduced to improve upon the Partial Dependency Plots method, which can introduce problems when features are correlated \cite{Molnar_2023}.

\[
\hat{f}_{S,ALE}(x_S) = \int_{z_{0,S}}^{x_S} \mathbb{E}_{X_C|X_S=x_S} \left[ \hat{f}^S (X_S, X_C) | X_S = z_S \right] dz_S - \text{constant}
\]

\[
= \int_{z_{0,S}}^{x_S} \left( \int_{X_C} \hat{f}^S (z_S, X_C) d\mathbb{P}(X_C|X_S = z_S) \right) dz_S - \text{constant}
\]

The ALE method calculates the differences in predictions by replacing the feature of interest with grid values \( z \). The difference in prediction is the effect the feature has for an individual instance in a certain interval. The sum on the right adds up the effects of all instances within an interval, which appears in the formula as the neighbourhood \( N_j(k) \). We divide this sum by the number of instances in this interval to obtain the average difference of the predictions for this interval. This average in the interval is what the term \emph{Local} in the name ALE covers. The sum symbol on the left means that we accumulate the average effects across all intervals. The uncentered ALE value of a feature that lies in the third interval, for example, is the sum of the effects of the first, second, and third intervals. This is what is reflected by the word \emph{Accumulated} in ALE.

This effect is centred so that the mean effect is zero.

\[
\hat{f}_{j,ALE}(x) = \hat{f}_{j,ALE}^{'}(x) - \frac{1}{n} \sum_{i=1}^{n} \hat{f}_{j,ALE}^{'}\left(x_{j}^{(i)}\right)
\]

\cite{Molnar_2023}\cite{apley2020visualizing}

\section{Related Works}

There are several papers and resources explore the implementation of QML algorithms, including guides provided through IBM Qiskit \cite{Qiskit2021} and papers such as  'Quantum variational multi-class classifier for the iris data set' \cite{piatrenka2022quantum}, This paper confirmed a method implemented to check optimising combinations of ansatzs and optimisers. However, there is only a handful of papers that explore the implementation of feature importance and explainability for QML.\\

There are many papers and resources about machine learning, feature importance, and quantum computing and quantum machine learning, however, there is an apparent gap in combining these topics. The first paper to compare feature importance between quantum and classical machine learning models was a paper titled 'Study of feature importance for quantum machine learning models' \cite{baughman2022study} was conducted by researchers at IBM using ESPN Fantasy Football data and claims to be the first paper exploring feature importance in QML. However, this paper only investigated two methods of feature importance; Permutation importance and ALE \cite{baughman2022study}. This article expands on this paper by providing a much more in-depth report on the implementation of QML and by both exploring more methods of feature importance and also venturing into methods of explainability. \\
There are fewer papers exploring the area of explainable QML; 'eXplainable AI for Quantum Machine Learning' \cite{steinmuller2022explainable} Implements a variation of SHAP on basic classifiers, ranging from a single qubit classifier two a four qubit classifier, using the binary 'bars and stripes' data. However, this paper does not make any comparisons to the results from a classical counterpart. \\ Last year, a paper published 'Explainable heart disease prediction using ensemble-quantum machine learning approach' \cite{abdulsalam2023explainable} utilised various QML methods on UCI's Heart Disease data set, with a binary classification of diagnosis or not. This paper implemented SHAP predictions but does not make clear comparisons between the classical and quantum results concerning their SHAP values or comparable metrics.

What this article does is implement classical and quantum machine learning models on the multi-class Iris dataset, which is a well-known data set used to classify flowers, and implement several methods of feature importance and explainability techniques and provide some direct comparisons between the results. 

\begin{table*}[!ht]
\centering
\caption{Literature Comparison}
\label{tab:literature_comparison}
\begin{tabular}{|p{4cm}|c|c|c|c|c|c|c|c|}
\hline
Paper & SVM/C & RF & VQC & QSVC & ALE & LOO & Perm & SHAP\\ \hline
Study of Feature Importance	for	Quantum	Machine	Learning Models. \cite{baughman2022study} & Yes & & Yes& &Yes & & Yes &  \\ \hline
Quantum variational multi-class classifier for the
iris data set. \cite{piatrenka2022quantum} & & &Yes & & & & & \\
\hline
Supervised learning with quantum-enhanced feature space. \cite{Havlicek2019} & Yes & &Yes &Yes & & & & \\
\hline
eXplainable AI for Quantum Machine Learning. \cite{steinmuller2022explainable} & Yes & &Yes & & & & & \\
\hline
Principles and practice of explainable machine learning. \cite{Belle_Papantonis_2021} & Yes & Yes& & & & & Yes & Yes\\
\hline
Explaining Quantum Circuits with Shapley Values (pre-print)
\cite{heese2023explaining} & & & Yes& Yes& & & &Yes\\
\hline
Explainable heart disease prediction using ensemble-quantum machine learning approach \cite{abdulsalam2023explainable}& & &Yes &Yes & & & &Yes \\ \hline
This article & Yes&Yes & Yes& Yes&Yes & Yes& Yes& Yes\\
\hline
\end{tabular}
\end{table*}

\section{Implementation}
We have simulated the quantum models with the Iris data set with Qiskit's SDK.

\subsection{The Iris Dataset}
The Iris dataset, introduced by Ronald Fisher in 1936, is a foundational dataset used in statistical learning and machine learning. It comprises 150 samples from three species of Iris flowers, each described by four features: sepal length, sepal width, petal length, and petal width. The dataset's simplicity yet capability to demonstrate the effectiveness of various algorithms has made it a staple for teaching purposes. And so this dataset was chosen to be used in this article. \cite{scikit_toy} \\
For a clear breakdown, of the Iris dataset, refer to the table. \ref{tab:iris_dataset}.

\begin{figure*}[h!]
  \centering
  \includegraphics[width=1\linewidth]{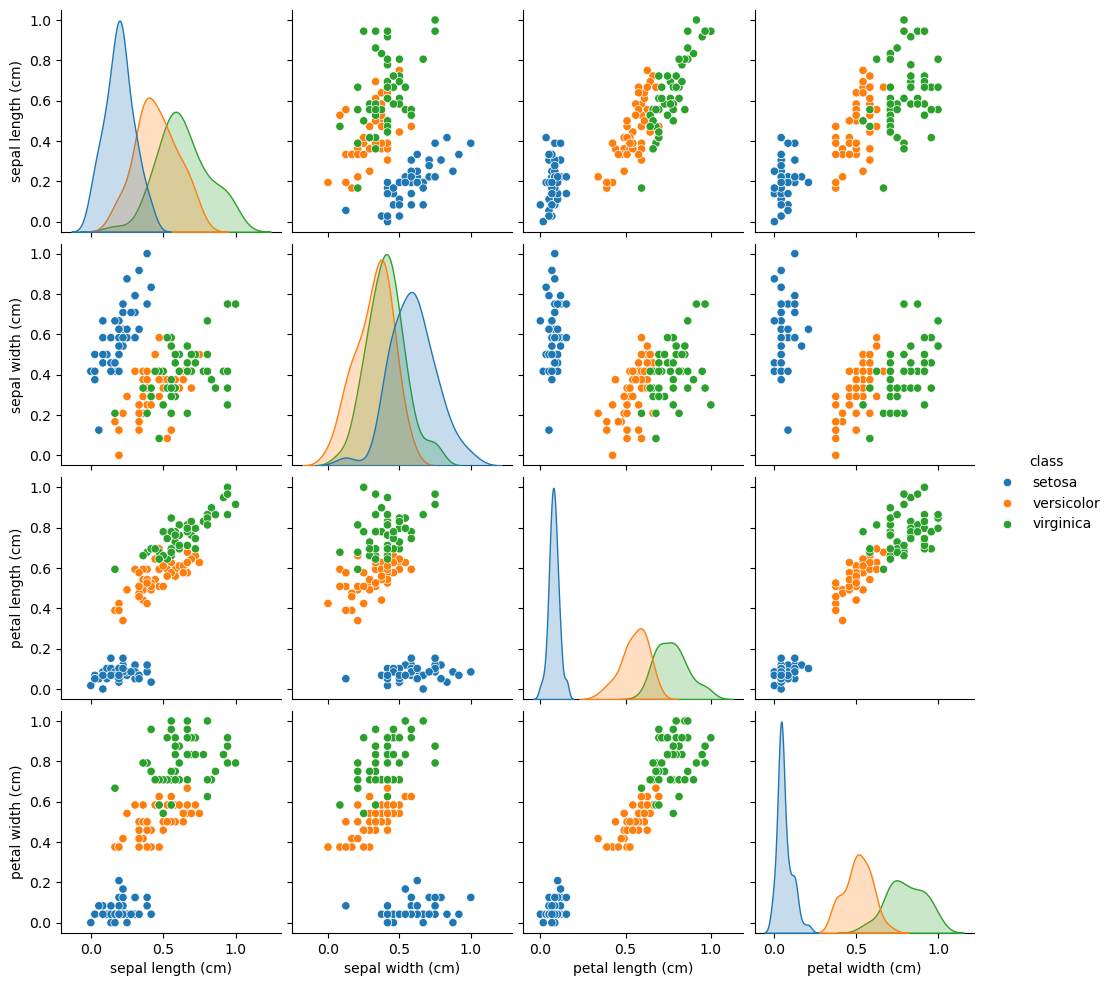} \caption[Iris pairs plot]{Iris Pairs Plot.}
  \label{fig:iris_pairs_plot}
\end{figure*}

\subsection{Data Preprocessing}
The data was normalised with scikit-learns MinMaxScaler, where features are scaled to fall from zero to one. This step was taken to ensure better comparability between the classical and quantum models, as this step was necessary to encode the data for the quantum classifiers.
All the models were trained and scored on an 80\% - 20\% train-test split. The breakdown of the test group can be found in table \ref{tab:test_cases}.

\begin{table}[hb!]
\centering
\caption{Test Case Distribution}
\label{tab:test_cases}
\begin{tabular}{@{}lcc@{}}
\toprule
\textbf{Species} & \textbf{Label} & \textbf{Test Count} \\ \midrule
Setosa       & 0     & 7          \\
Versicolor   & 1     & 9          \\
Virginica    & 2     & 14         \\ \bottomrule
\end{tabular}
\end{table}

\begin{table}[h]
\centering
\caption{Iris Dataset Summary}
\label{tab:iris_dataset}
\begin{tabular}{|l|l|}
\hline
\textbf{Characteristic} & \textbf{Detail} \\
\hline
Number of Instances & 150 (50 in each of three classes) \\
\hline
Number of Attributes & 4 numeric, predictive attributes and the class \\
\hline
Attribute Information & Sepal length in cm \\
                      & Sepal width in cm \\
                      & Petal length in cm \\
                      & Petal width in cm \\
\hline
Classes & Iris-Setosa \\
        & Iris-Versicolour \\
        & Iris-Virginica \\
\hline
Missing Attribute Values & None \\
\hline
Class Distribution & 33.3\% for each of 3 classes \\
\hline
Creator & R.A. Fisher \\
\hline
Donor & Michael Marshall \\  &(MARSHALL\%PLU@io.arc.nasa.gov) \\
\hline
Date & July, 1988 \\
\hline
\end{tabular}
\end{table}

Plot \ref{fig:iris_pairs_plot}  shows a pairs plot of the Iris dataset, which visualises the relationship between the features. The consistent separation of the setosa data points suggests that it will be more easily classified than versicolor and virginica, which have several overlapping points. It is also worth noting the linear structure between peal length and width which shoes the correlation between the two features, which can indicate that one of the features may be redundant.

\subsubsection{VQC Implementation}
The first and most basic QML model implemented was the VQC, which could be described as the quantum equivalent of a simple linear classifier. A ZZFeatureMap was used on the dataset to use the classical data in the quantum algorithm. However, to maximise the performance of the model we also needed to select what combination of ansatz and classical optimizer should be utilised.\\

\subsubsection*{VQC Optimisations}

\textbf{Choices of Ansatz} \\
\begin{itemize}
    \item \textbf{RealAmplitudes} circuit is a heuristic trial wave function used as Ansatz in chemistry applications or classification circuits in machine learning. The circuit consists of alternating layers of Y rotations and CX entanglements. The entanglement pattern can be user-defined or selected from a predefined set. It is called RealAmplitudes since the prepared quantum states will only have real amplitudes, the complex part is always 0. \cite{RealAmp}
    \item \textbf{EfficientSU2} circuit consists of layers of single qubit operations spanned by SU(2) and  CX entanglements. This is a heuristic pattern that can be used to prepare trial wave functions for variational quantum algorithms or classification circuit for machine learning. \\ SU(2) stands for special unitary group of degree 2, its elements are 2 × 2
 unitary matrices with determinant 1, such as the Pauli rotation gates. \cite{EffSU2}
\end{itemize}

\textbf{Choices of Optimizer} \\
\begin{itemize}
    \item \textbf{COBYLA} is a numerical optimization method for constrained problems where the derivative of the objective function is not known. \cite{COBYLA}
    \item \textbf{SLSQP} minimizes a function of several variables with any combination of bounds, equality and inequality constraints. SLSQP is ideal for mathematical problems for which the objective function and the constraints are twice continuously differentiable. \cite{SLSQP}

\end{itemize}

\begin{figure*}[h!]
  \centering
  \includegraphics[width=1\linewidth]{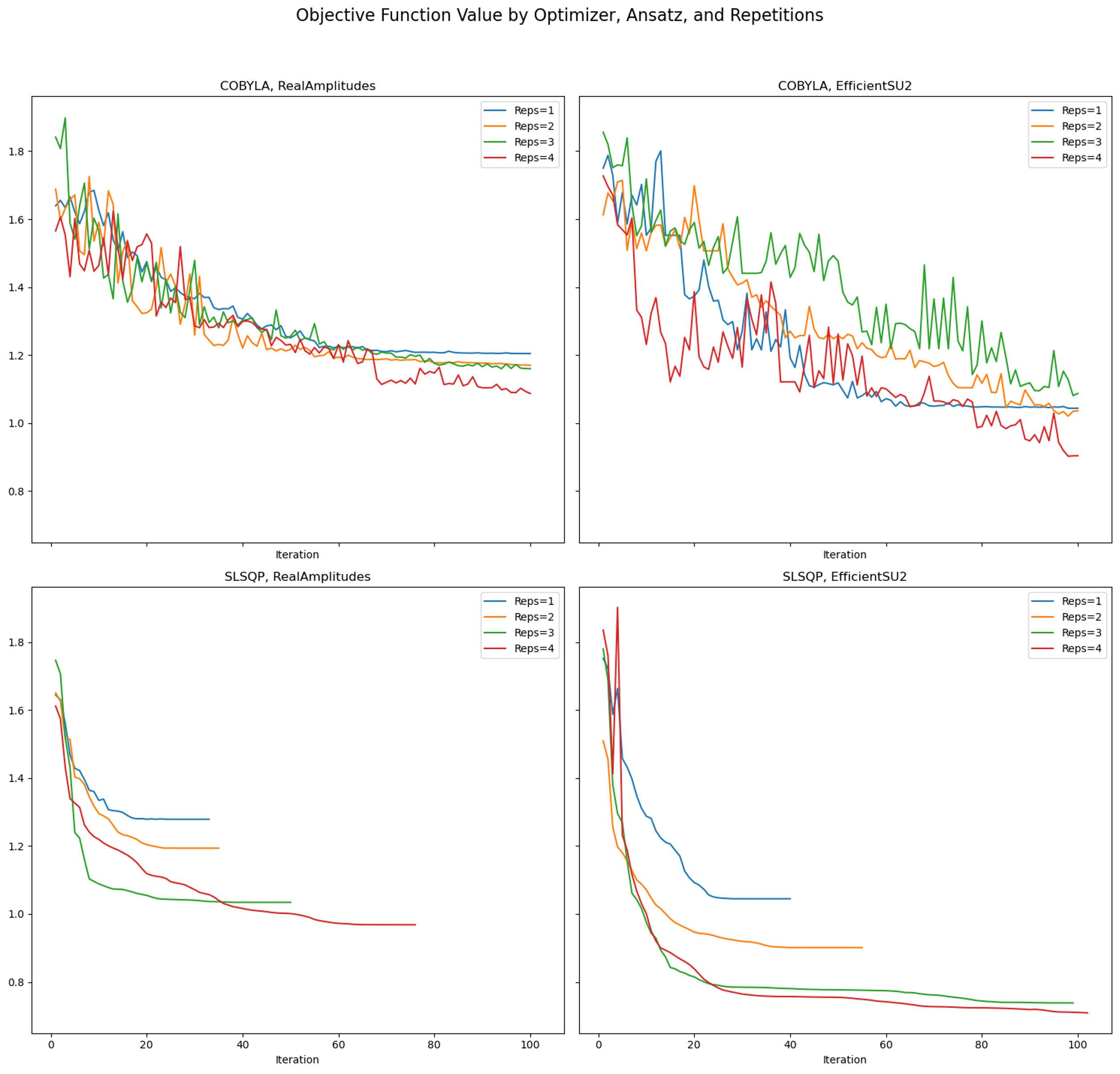} \caption[VQC Optimisation]{Combinations of Optimiser and Ansatz for the VQC.}
  \label{fig:iris_combo_optimal}
\end{figure*}

These were combined and iterated from one to four repetitions of the ansatz. As seen in \ref{fig:iris_combo_optimal}, SLSQP and EfficentSU2 have the best overall performance, with an initial score of 90\%, the SLSQP and EfficientSU2 with three and four reps would take between one and one and a half hours respectively to train on a local simulator, while the COBYLA and EfficientSU2 with three reps only took a minute and a half, with a bootstrapped accuracy of 87\%. Due to the necessity of high numbers of repetitions in the experiments, this model was selected for the experiments.

\subsubsection{QSVC Implementation}
The Quantum Support Vector Classifier is the other quantum model implemented in this article, which is the quantum equivalent of the Support Vector Classifier. To build the QSVC, we need to specify a quantum kernel. The FidelityQuantumKernel class in Qiskit provides a quantum kernel implementation based on the principle of state fidelity. This takes in a feature map, for which we again use the ZZFeatureMap and a fidelity instance; ComputeUncompute which uses the sampler primitive to calculate the state fidelity of two quantum circuits following the compute-uncompute method. \cite{kernel_QML}

\section{Results}
Table \ref{tab:model_summary} provides the classification report generated from each of the models; the SVC showcases high precision across all classes, achieving perfect precision and recall for class 0 (setosa). Its performance remains strong for classes 1 (Versicolor) and 2 (virginica), with high f1-scores and overall accuracy of 93\%. \\
The VQC had decent precision for classes 1 and 2, with scores of 90\% and 92\% respectively, with a high f1-score for class 1, but across almost all metrics, VQC appears to be the weakest classifier.\\
The QSVC presents excellent precision and perfect recall for classes 0 and 2, and a recall of 86\% for class 1. The f1-scores are notably high, particularly for class 2, with an overall accuracy reaching 97\%. This model demonstrates the best performance among the other classifiers, indicating its possibly superior capability to manage and balance dataset complexities.\\
The Random Forest classifier model exhibits 100\% precision for classes 0 and 2 but slightly lower precision for class 1. While it achieves perfect recall for class 0 and class 1, its recall for class 2 drops, reflecting some difficulties in consistently identifying this class. The f1-scores are high for class 0 but moderate for classes 1 and 2, culminating in an overall accuracy of 90\%. Although the model performs well, it shows specific weaknesses that could be mitigated through further optimization.\\
Overall, the QSVC stands out as the superior model, offering the highest accuracy and robust performance across all evaluated metrics, making it the preferred choice when quantum resources are available. The SVC follows closely behind, providing a highly effective and reliable alternative without the need for quantum enhancements. Both the SVC and QSVC exhibit strong capabilities in handling the classification challenges presented by the Iris dataset, whereas the RF, despite its strengths, shows areas for potential improvement.

\begin{figure*}[h!]
  \includegraphics[width = 1\linewidth]{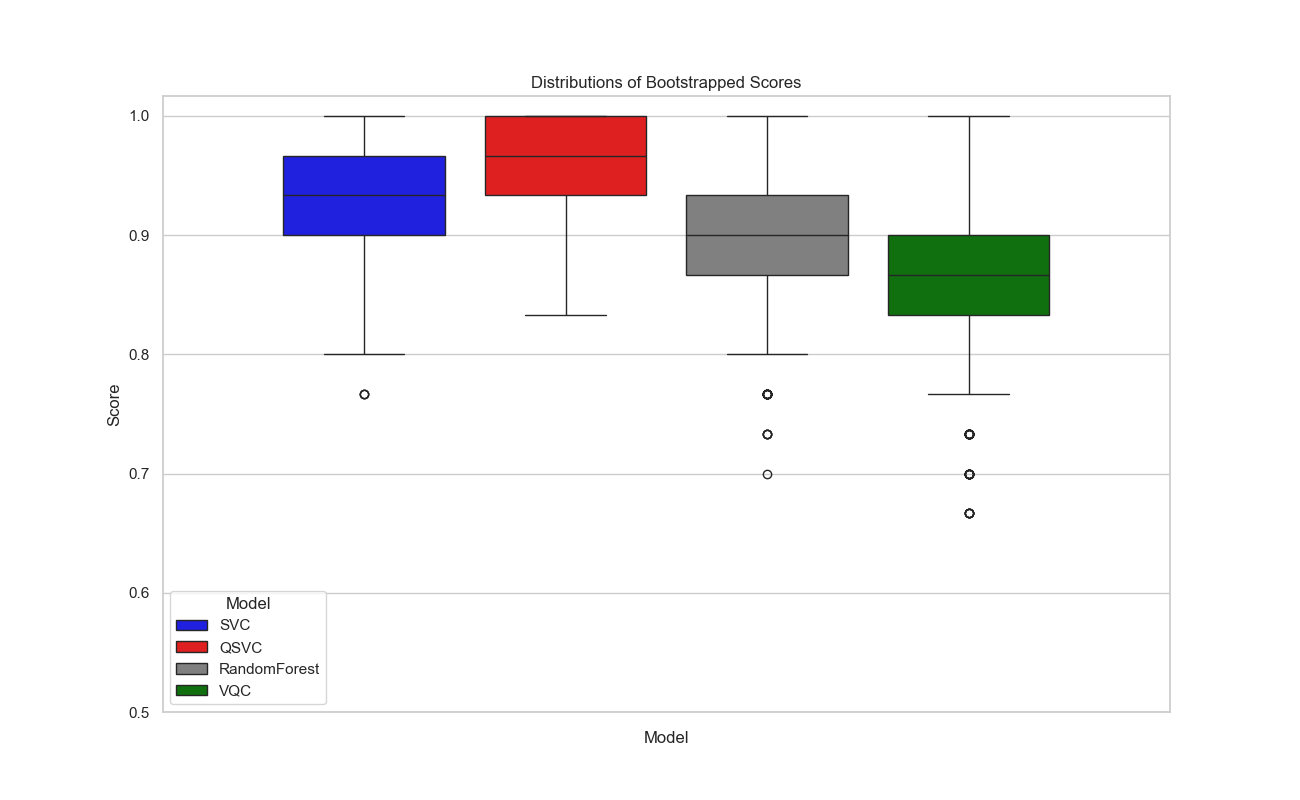} \caption[Model Accuracy Comparison]{Comparison of bootstrapped accuracy scores.}
  \label{fig:Model_accuracy_boxplot_comparison}
\end{figure*}

To compare the models' overall accuracy against the test data, a bootstrap implementation was used using 1000 bootstrap resamples from the test data, as seen in table \ref{tab:bootstrap_accuracy}. The QSVC and SVC had average bootstrap results of 97\% and 93\%  respectively, outperforming the random forest classifier and VQC. The box-plot \ref{fig:Model_accuracy_boxplot_comparison} shows how the distributions of the results overlap, which can indicate that these models could be suitable for comparison. 

\begin{table}[ht]
\centering
\caption{Summary of Classification Results by Model}
\label{tab:model_summary}
\begin{tabular}{@{}llllll@{}}
\toprule
\textbf{Model} & \textbf{Metric} & \textbf{Class 0} & \textbf{Class 1} & \textbf{Class 2} & \textbf{Overall} \\ \midrule
\multirow{4}{*}{VQC} 
 & Precision & 0.90 & 0.71 & 0.92 & \\
 & Recall    & 1.00 & 0.71 & 0.86 & \\
 & F1-Score  & 0.95 & 0.71 & 0.89 & \\
 & Accuracy  &      &      &      & 0.87 \\ \midrule
\multirow{4}{*}{SVC} 
 & Precision & 1.00 & 0.78 & 1.00 & \\
 & Recall    & 1.00 & 1.00 & 0.86 & \\
 & F1-Score  & 1.00 & 0.88 & 0.92 & \\
 & Accuracy  &      &      &      & 0.93 \\ \midrule
\multirow{4}{*}{QSVC}
 & Precision & 1.00 & 1.00 & 0.93 & \\
 & Recall    & 1.00 & 0.86 & 1.00 & \\
 & F1-Score  & 1.00 & 0.92 & 0.97 & \\
 & Accuracy  &      &      &      & 0.97 \\ \midrule
\multirow{4}{*}{RF}  
 & Precision & 1.00 & 0.70 & 1.00 & \\
 & Recall    & 1.00 & 1.00 & 0.79 & \\
 & F1-Score  & 1.00 & 0.82 & 0.88 & \\
 & Accuracy  &      &      &      & 0.90 \\ \bottomrule
\end{tabular}
\end{table}

\begin{table}[htbp]
\centering
\caption{Bootstrapped Accuracy Results}
\label{tab:bootstrap_accuracy}
\begin{tabular}{@{}lccc@{}}
\toprule
Classifier    & Mean Accuracy & 25th Percentile & 75th Percentile \\ \midrule
SVC           & 0.932233      & 0.900000        & 0.966667        \\
QSVC          & 0.965567      & 0.933333        & 1.000000        \\
Random Forest & 0.900333      & 0.866667        & 0.933333        \\
VQC           & 0.864833      & 0.833333        & 0.900000        \\ \bottomrule
\end{tabular}
\end{table}

\begin{figure*}[h!]
  \includegraphics[width = 1\linewidth]{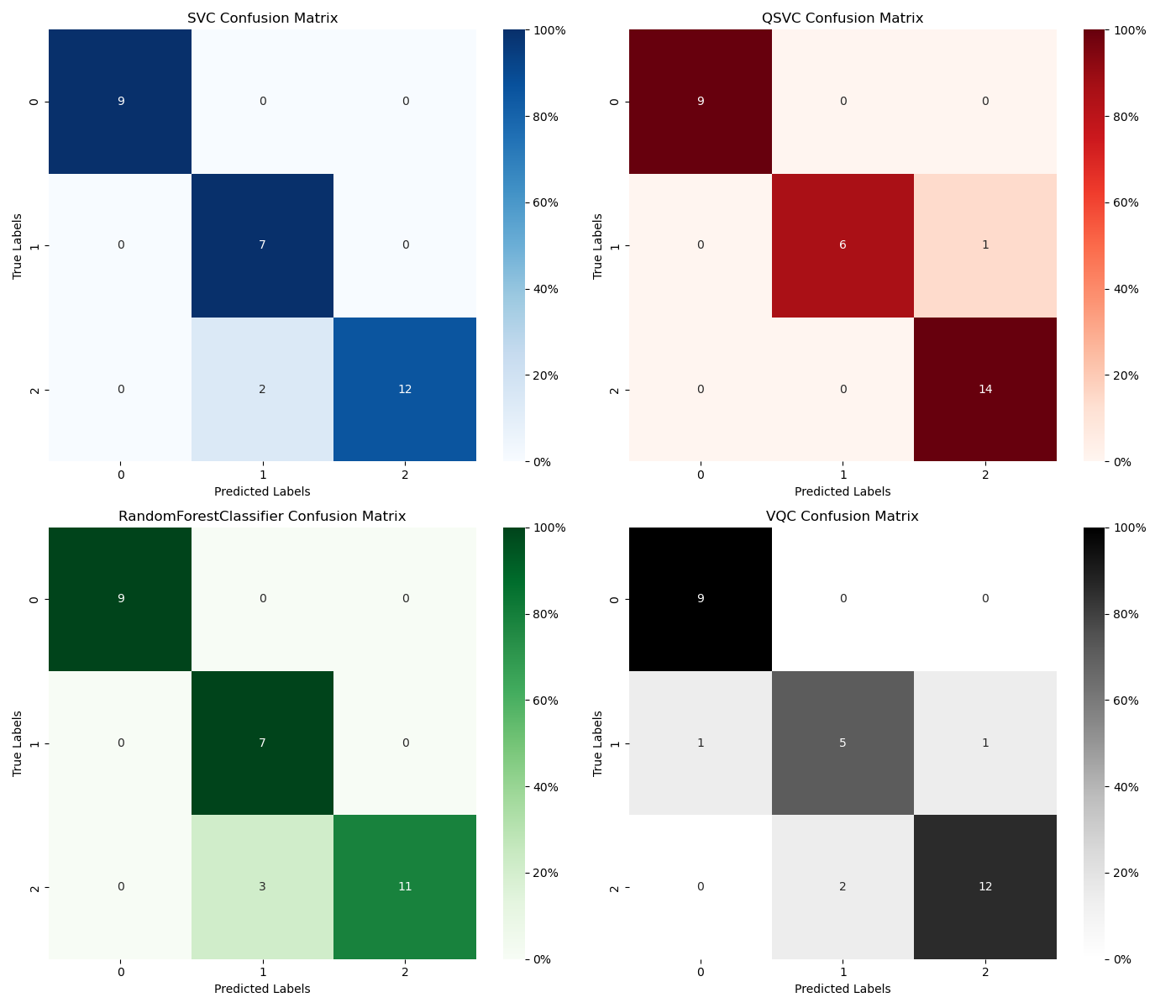} \caption[Confusion Matrices]{Confusion matrices of models.}
  \label{fig:confusion_matrices}
\end{figure*}

For a clearer idea of how the models performed with respect to the individual test case classes, we can look at the confusion matrix in \ref{fig:confusion_matrices}, which shows how the four models accurately classified all the test setosa (class 0) samples. This was expected and was also a handy sanity check as the setosa flower is clearly separable from the other two, which can be seen in the pairs plot \ref{fig:iris_pairs_plot}. The most interesting observation is that the QSVC could accurately classify all 14 test instances of virginica (class 2). The QSVC also misclassified one of the versicolor (class 1)  samples as virginica while both the SVC and Random Forest correctly classified all seven samples. The VQC had the poorest results by comparison, misclassified two of the versicolor samples, one as setosa and one as virginica.
A breakdown of the misclassification can be found in table \ref{tab:misclassification}.

\begin{table}[h]
\centering
\caption{Misclassification Details}
\label{tab:misclassification}
\begin{tabular}{c p{1.5cm} p{4cm}}
\toprule
\textbf{Test Sample} & \textbf{Label} &  \textbf{Misclassified As} \\
\midrule
4 & 2 & SVC: 1, Random Forest: 1, VQC: 1 \\
10 & 1 & QSVC: 2, VQC: 2 \\
15 & 1 & VQC: 0 \\
16 & 2 & Random Forest: 1 \\
17 & 2 & SVC: 1, Random Forest: 1, VQC: 1 \\
\bottomrule
\end{tabular}
\end{table}

\begin{figure*}[h!]
  \includegraphics[width = 1\linewidth]{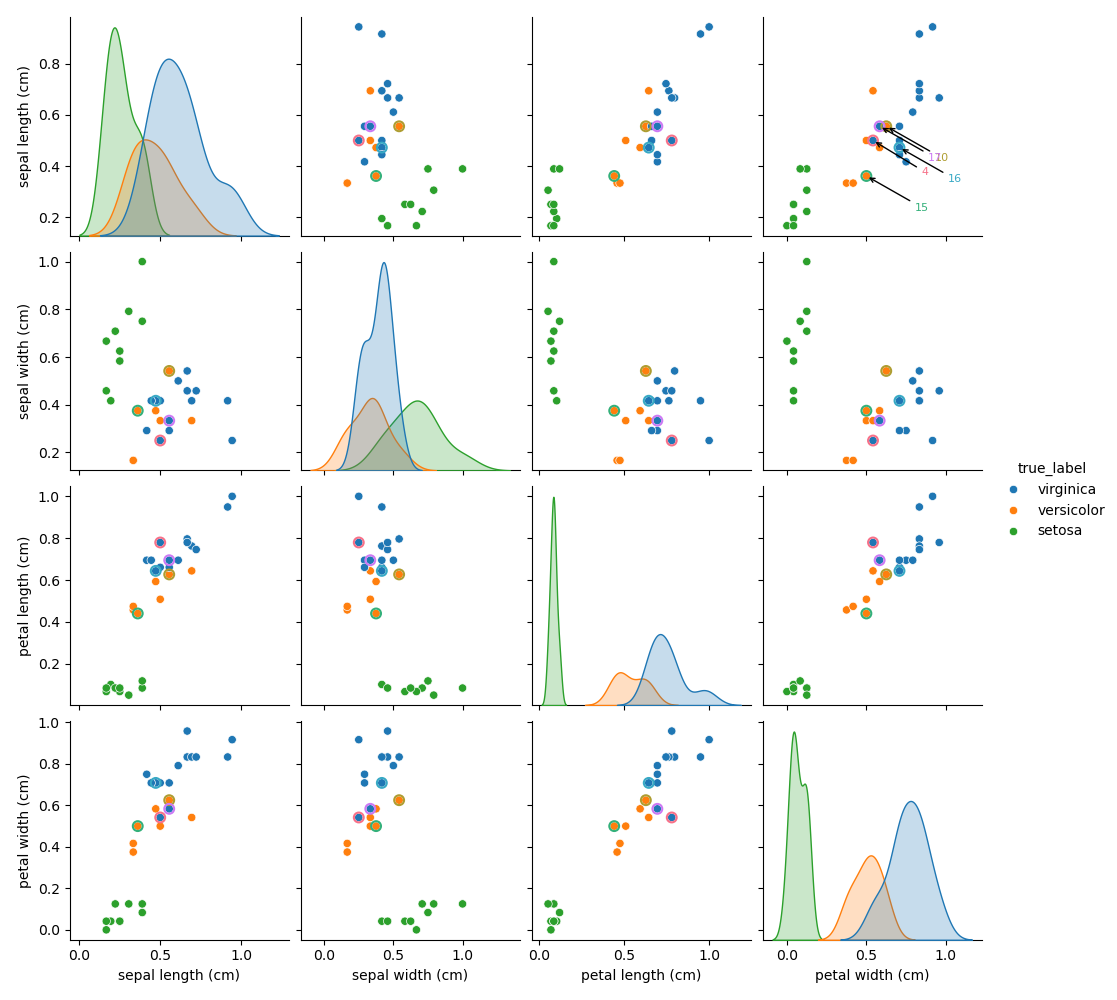} \caption[Test Points]{Pairs plot of the test data points, highlighting the misclassified points. For misclassification breakdown refer to table \ref{tab:misclassification}}
  \label{fig:test_pair_plot}
\end{figure*}

From table \ref{tab:misclassification} we can see that there are five of the thirty test cases were misclassified, with sample 4, which has a true label of 2 (virginica) misclassified as 1 (versicolor) by all models except the QSVC. For a view of what points were misclassified, refer to the plot \ref{fig:test_pair_plot}, which highlights the five misclassified points. We can see that four of the points occur where there are some overlaps of the classes. We can see in this plot that point 15 (highlighted in green), which was misclassified by the VQC, seems to be clearly grouped with other versicolor samples. This suggests that there is still further optimisation that could be implemented in the VQC.

\section{Result Analysis and Discussion} 
\subsection{Feature Importance}
Three methods of feature importance were implemented; Leave One Out, Permutation Importance, and Accumulated Local Effects. We will compare the results on a model by model basis, and an overall summary of each method results.

\subsubsection{Leave One Out}
To calculate LOO feature importance, we rerun the model omitting the feature of interest, and check how the performance of the model is affected. \\

\subsubsection*{SVC Leave-One-Out}

\begin{figure}[h!]
  \centering
  \includegraphics[width=1\linewidth]{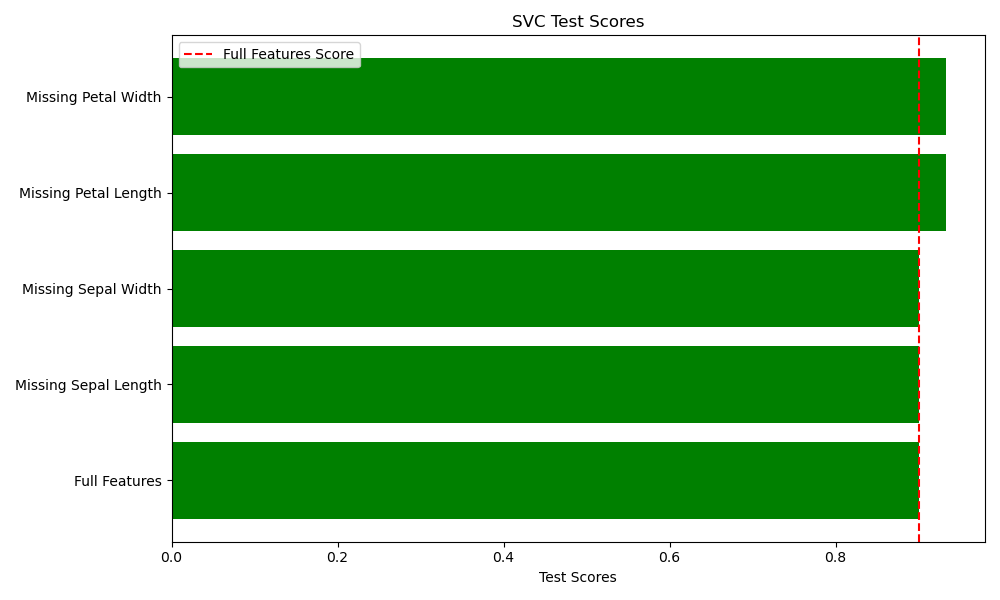} \caption[LOO Feature Importance]{SVC Leave One Out Feature Importance.}
  \label{fig:LOO_SVC}
\end{figure}

The effects of the feature omissions in the SVC can be seen in the plot \ref{fig:LOO_SVC}, and in the table \ref{tab:loo_svc}. We can see from the results that when either petal length or petal width is missing, the accuracy of the model increases by 3.3\%. This may indicate that petal length or petal length could be poor predictors for the SVC. This counterintuitive increase in test scores upon removing 'Petal Width' and 'Petal Length' could indicate an over-reliance on these features or it could be the removal of an interference effect due to the high correlation between these two feature, suggesting that only one of them are really needed in the model.  Accuracy remains unchanged when the sepal length or width is omitted. This suggests that the sepal length or width may not be a critical feature for the model, possibly because other features provide similar or sufficient information for making accurate classifications.\\
Due to the minimal differences, this analysis suggests that SVC regards all of the features to a similar importance.

\begin{table}[ht]
\centering
\caption{SVC LOO Results}
\label{tab:loo_svc}
\begin{tabular}{@{}p{3cm}p{3cm}@{}}
\toprule
\textbf{Scenario} & \textbf{Accuracy Score} \\
\midrule
Full Feature & 0.90 \\
Missing Sepal Length & 0.90 \\
Missing Sepal Width & 0.90 \\
Missing Petal Length & 0.93 \\
Missing Petal Width & 0.93 \\
\bottomrule
\end{tabular}
\end{table}

\subsubsection*{Random Forest Leave-One-Out}

\begin{figure}[h!]
  \centering
  \includegraphics[width=1\linewidth]{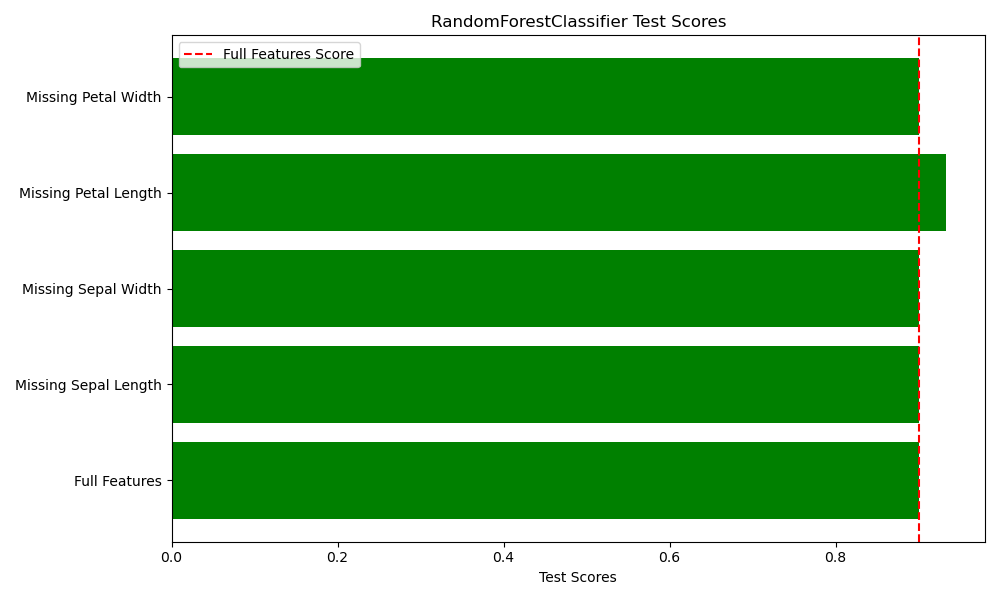} \caption[RF Feature Importance]{Random Forest Leave One Out Feature Importance.}
  \label{fig:LOO_rf}
\end{figure}

The effects of the feature omissions in the Random Forest can be seen in the plot \ref{fig:LOO_vqc}, and in the table \ref{tab:loo_vqc}. We can see from the results that when either petal length is missing, the accuracy of the model increases by 3.3\%. This may indicate that petal length could be a poor predictor compared to the others. This is interesting due to the high correlation between petal length and petal width, which would suggest that if petal length has an effect, petal width would cause a similar effect. Accuracy remains unchanged when the sepal length or width is omitted. This suggests that the sepal length or width may not be a critical feature for the model, possibly because other features provide similar or sufficient information for making accurate classifications.\\
Due to the minimal differences, this analysis suggests that the Random Forest regards all of the features to similar importance like the SVC possibly assigning petal length with the lowest importance. 

\begin{table}[ht]
\centering
\caption{Random Forest LOO Results}
\label{tab:loo_vqc}
\begin{tabular}{@{}p{3cm}p{3cm}@{}}
\toprule
\textbf{Scenario} & \textbf{Accuracy Score} \\
\midrule
Full Feature & 0.90 \\
Missing Sepal Length & 0.90 \\
Missing Sepal Width & 0.90 \\
Missing Petal Length & 0.93 \\
Missing Petal Width & 0.90 \\
\bottomrule
\end{tabular}
\end{table}

\subsubsection*{VQC Leave-One-Out}

\begin{figure}[h!]
  \centering
  \includegraphics[width=1\linewidth]{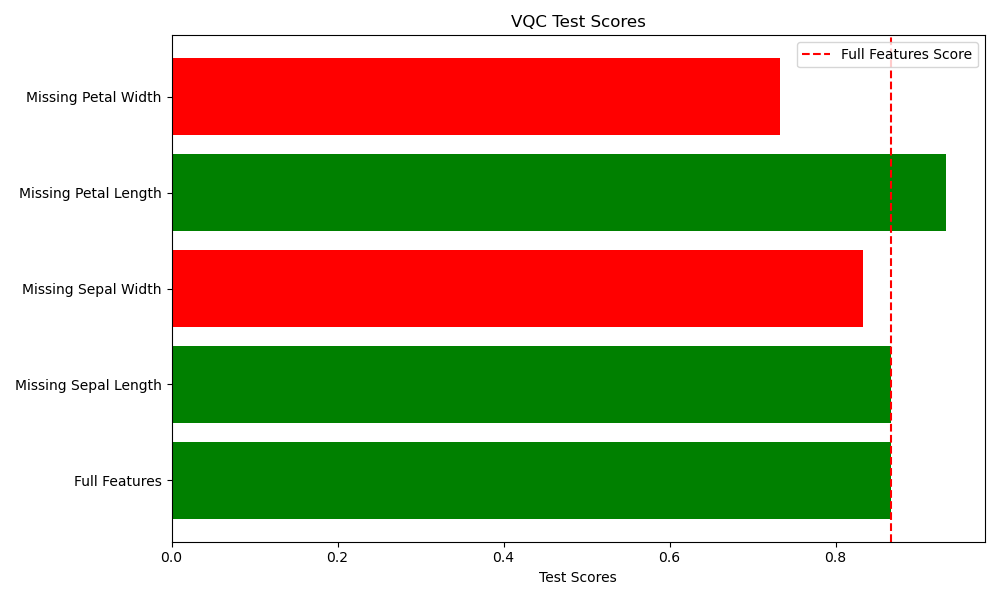} \caption[VQC Feature Importance]{VQC Leave One Out Feature Importance.}
  \label{fig:LOO_vqc}
\end{figure}
The effects of the feature omissions in the VQC can be seen in the plot \ref{fig:LOO_vqc}, and in the table \ref{tab:loo_rf}. The LOO results are much different to the SVC and Random Forest. The full-featured model has an accuracy of 87\%, but this time the removal of petal width caused a drop to 73\%, a 16\% decrease from our full model, a much bigger difference to the 3.3\% difference we found with the SVC and no change in the Random Forest classifier. This causes the effect of removing petal length to be somewhat paradoxical, increasing the model's performance by 6.9\%. Due to the highly correlated nature of petal length and width, they would be expected to have similar predictive effects in a model, however, they have opposite effects on the model's performance. \\
We can also see that the effect of removing sepal width, which had no effect in either the SVC or Random Forest, caused a 4.6\% decrease in the VQC's accuracy. These results demonstrate that the VQC has highly different levels of feature importance, with petal width being considered the most important, as its absence causes the biggest decrease in accuracy, followed by sepal width for the same reason, then sepal length and petal length being the least important as its absence improves the models' accuracy. 

\begin{table}[ht]
\centering
\caption{VQC LOO Results}
\label{tab:loo_rf}
\begin{tabular}{@{}p{3cm}p{3cm}@{}}
\toprule
\textbf{Scenario} & \textbf{Accuracy Score} \\
\midrule
Full Feature & 0.87 \\
Missing Sepal Length & 0.87 \\
Missing Sepal Width & 0.83 \\
Missing Petal Length & 0.93 \\
Missing Petal Width & 0.73 \\
\bottomrule
\end{tabular}
\end{table}

\subsubsection*{QSVC Leave-One-Out}

\begin{figure}[h!]
  \centering
  \includegraphics[width=1\linewidth]{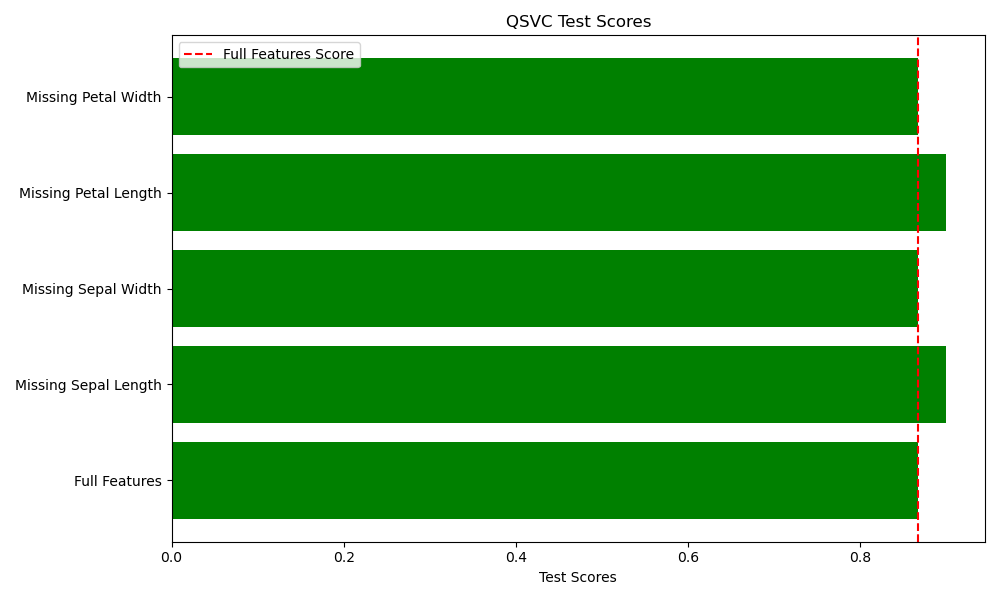} \caption[QSVC Feature Importance]{QSVC Leave One Out Feature Importance.}
  \label{fig:LOO_qsvc}
\end{figure}

The effects of the feature omissions in the VQC can be seen in the plot \ref{fig:LOO_qsvc}, and in the table \ref{tab:loo_qsvc}. We can see that the QSVC had similar accuracy to the VQC, but ended up with highly different results, more in line with what we observed in the SVC and Random Forest, the removal of individual features had very little overall effect on the model's accuracy. What has remained consistent across all the results was the removal of petal length slightly increased the model's accuracy, in this case by 3.5\%. There was also an identical effect with the removal of sepal length, indicating that sepal length introduces some noise into the model.\\
According to this LOO analysis, we can see that QSVC acts somewhat similarly to the classical models, regarding all the models with similar importance.

\begin{table}[ht]
\centering
\caption{QSVC LOO Results}
\label{tab:loo_qsvc}
\begin{tabular}{@{}p{3cm}p{3cm}@{}}
\toprule
\textbf{Scenario} & \textbf{Accuracy Score} \\
\midrule
Full Feature & 0.87 \\
Missing Sepal Length & 0.90 \\
Missing Sepal Width & 0.87 \\
Missing Petal Length & 0.90 \\
Missing Petal Width & 0.87 \\
\bottomrule
\end{tabular}
\end{table}

\subsubsection*{LOO Summary}
\begin{figure*}[h!]
  \includegraphics[width = 1\linewidth]{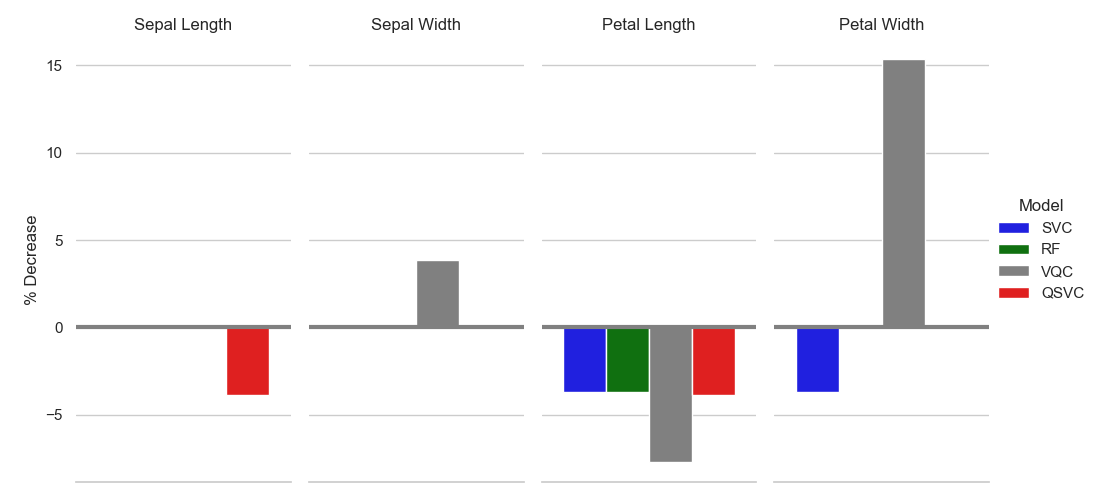} \caption[Overall LOO]{Overall LOO results.}
  \label{fig:loo_comp}
\end{figure*}

Figure \ref{fig:loo_comp}, shows a full comparison of LOO feature importance across all the models. The VQC model exhibits a highly different approach to assessing feature importance, particularly diverging from the other models in its valuation of Petal Width. The VQC was also the only model to demonstrate a decrease in performance at the removal of any features. The removal of a feature in the other three models either did not affect accuracy or increased it. Interestingly, the QSVC, while also a quantum model, seems more aligned with the classical SVC and RF models in terms of feature impact, except with Sepal width, whose removal increases accuracy. Both quantum models demonstrate a sensitivity to Sepal features, yet they differ in which specific Sepal features they find more important. The QSVC shows a marked increase in importance for Sepal Length, a trend not observed in the VQC model. Meanwhile, the VQC assigns considerable importance to Sepal Width, unlike the QSVC. This difference demonstrates the unintuitive ways in which quantum models process feature importance compared to the classical models.

\subsubsection{Permutation Importance}
Permutation importance differs from LOO by instead of removing the feature entirely, the feature values will be repeatedly shuffled to break the relationship between the feature and the class. This method will allow us to get a gauge of the variation of the importance.

\subsubsection*{SVC Permutation Importance}
\label{section:svc_perm}
\begin{figure}[h!]
  \centering
  \includegraphics[width=1\linewidth]{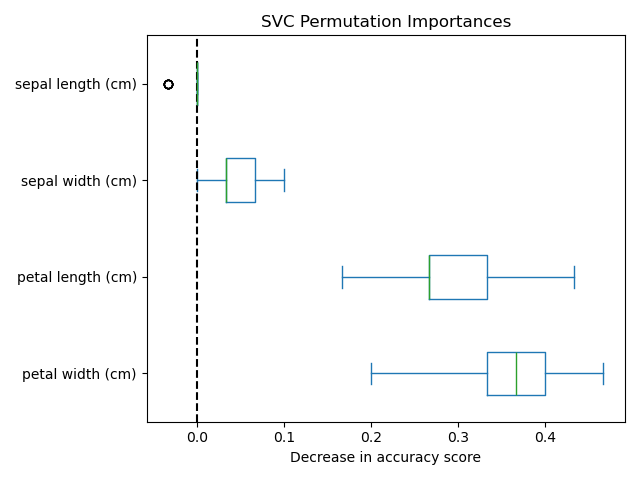} \caption[Permutation SVC Feature Importance]{SVC Permutation Importance.}
  \label{fig:perm_SVC}
\end{figure}

We can see in the plot \ref{fig:perm_SVC} that petal width and petal length have the biggest impact when their values are permuted, and sepal width seems to have little to no effect on the model accuracy. The permutation of sepal length has no effect whatsoever. The results of petal length and petal width being permutated are extremely interesting as they appear to demonstrate the opposite effect in the LOO analysis, seen in the plot \ref{fig:LOO_SVC}. This result makes more intuitive sense as petal length and width are considered to have strong predictive power for the Iris dataset. However, this may indicate that the model may be better off with just petal width rather than both petal width and petal length. Another possible cause is the Permutation Importance artificially introduces noise in the data so that it would be highly unlikely that a model's performance would increase when a feature is permuted. 

\subsubsection*{Random Forest Permutation Importance}

\begin{figure}[h!]
  \centering
  \includegraphics[width=1\linewidth]{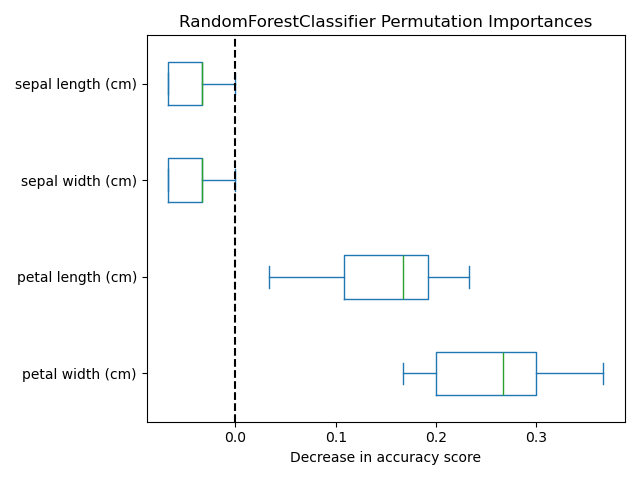} \caption[Permutation RF Feature Importance]{Random Forest Permutation Importance.}
  \label{fig:perm_RF}
\end{figure}

We can see in the plot \ref{fig:perm_RF} that petal width and petal length have the biggest impact when their values are permuted, while sepal length and sepal width seem to marginally increase the model's accuracy. These results are also quite different compared to the LOO analysis in plot \ref{fig:LOO_rf}, where the removal of petal length causes a slight increase in the model's accuracy. This result closely aligns with the SVC permutation importance results, which also follow the intuition of petal length and width being strong predictors. We could assume the same reasons mentioned in the SVC Permutation analysis \ref{section:svc_perm} for the difference seen in this result and the LOO.

\subsubsection*{VQC Permutation Importance}

\begin{figure}[h!]
  \centering
  \includegraphics[width=1\linewidth]{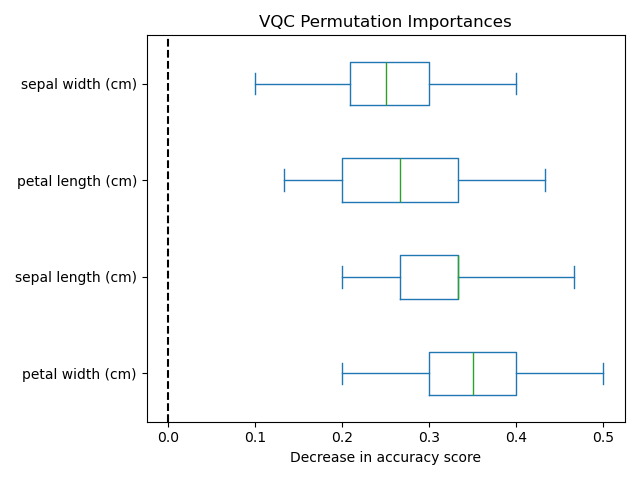} \caption[Permutation VQC Feature Importance]{VQC Permutation Importance.}
  \label{fig:perm_vqc}
\end{figure}

The results of Permutation Importance, in plot \ref{fig:perm_vqc} shows that the VQC has a much more balanced distribution of the features' importance, with sepal length having the second highest importance after petal width, which differs from the two classical models who both had a very similar ranking of petal width and length being the most importance with sepal length and width having little to none. The size of the whiskers on the plots and the high interquartile ranges further show how similarly the VQC considers the features. The permutation results still differ from the LOO analysis. Where the removal of petal length increased model performance, and removing sepal length did not affect performance.

\subsubsection*{QSVC Permutation Importance}

\begin{figure}[h!]
  \centering
  \includegraphics[width=1\linewidth]{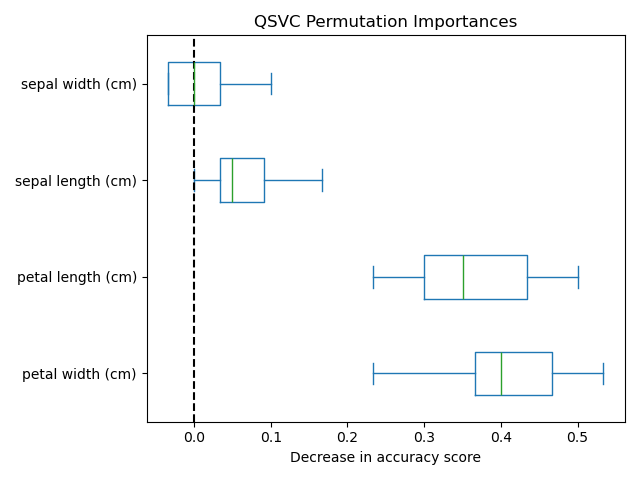} \caption[Permutation QSVC Feature Importance]{QSVC Permutation Importance.}
  \label{fig:perm_qsvc}
\end{figure}

We can see that again in plot \ref{fig:perm_qsvc} that the QSVC model follows a similar distribution of feature importance to the classical models, with petal length and petal width causing the biggest decrease in model accuracy when their values are permuted, and sepal width and sepal length causing slightly bigger effects but still much lower than the petal features. We can see that the LOO results demonstrated no effects with the removal of petal width and sepal width, and removing petal length and sepal length causes a slight increase in model accuracy. 

\subsubsection*{Permutation Importance Summary}

\begin{figure*}[h!]
  \includegraphics[width=1\linewidth]{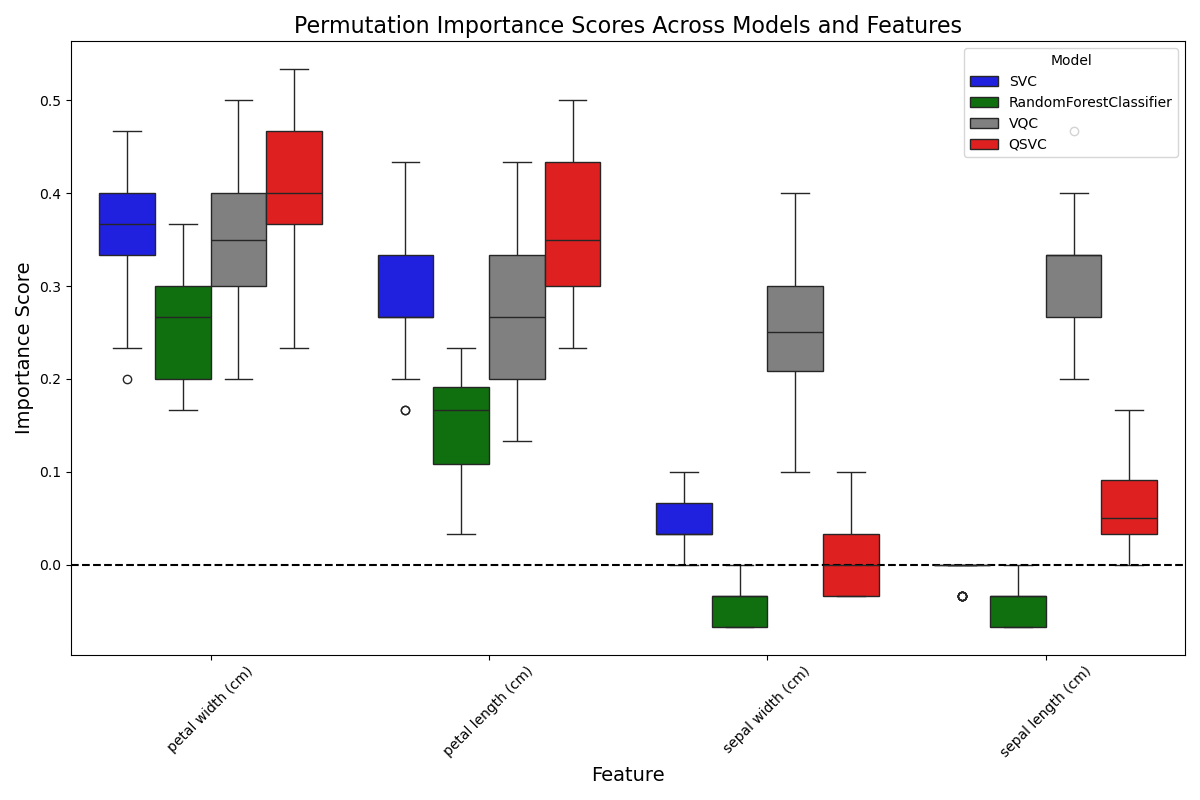} \caption[Permutation Overall Feature Importance]{Overall Permutation Importance.}
  \label{fig:perm_combo}
\end{figure*}

The plot \ref{fig:perm_combo} we can see that the VQC has a fundamentally different approach to feature importance compared to all the other models, including the QSVC, with almost equal weighting across all the features. For the other three models, we can a clear preference for petal length and petal width over sepal length and sepal width, with the lower wicks of the petal features being above the top wicks of the sepal features. The QSVC does seem to place slightly more importance on the sepal features than the classical models, but one should note that depending on the permutation of sepal width, the median effect was zero, with some permutations resulting in a better performance, but tending to worse, indicated by the high whisker. \\
Overall we can see that according to permutation importance analysis, petal features have a higher ranking importance than sepal. This seems to follow LOO results, which shows that petal features have a larger effect than the sepal counterparts, but seemingly in an opposite direction (\ref{fig:loo_comp}).

\subsubsection{Accumulated Local Effects Feature Importance}
\label{section: ALE_FI}
ALE computes feature importance differently to the LOO and Permutation Importance methods, rather than using model accuracy, ALE uses probabilities to measure how much the probability of making a classification changes concerning how the value of the feature changes. This can be aggregated and used to get an overall score of a model's feature importance, which will be analysed briefly.
Due to the probabilistic nature of ALE, the VQC model is unable to produce a probability vector, as this functionality was deprecated with Qiskit 1.0 \cite{IBM_deprecated}. Thus the QSVC model was used as the quantum comparison.  For a clearer comparison, we will compare the SVC against its quantum counterpart. 

\begin{figure*}[h!]
  \includegraphics[width=1\linewidth]{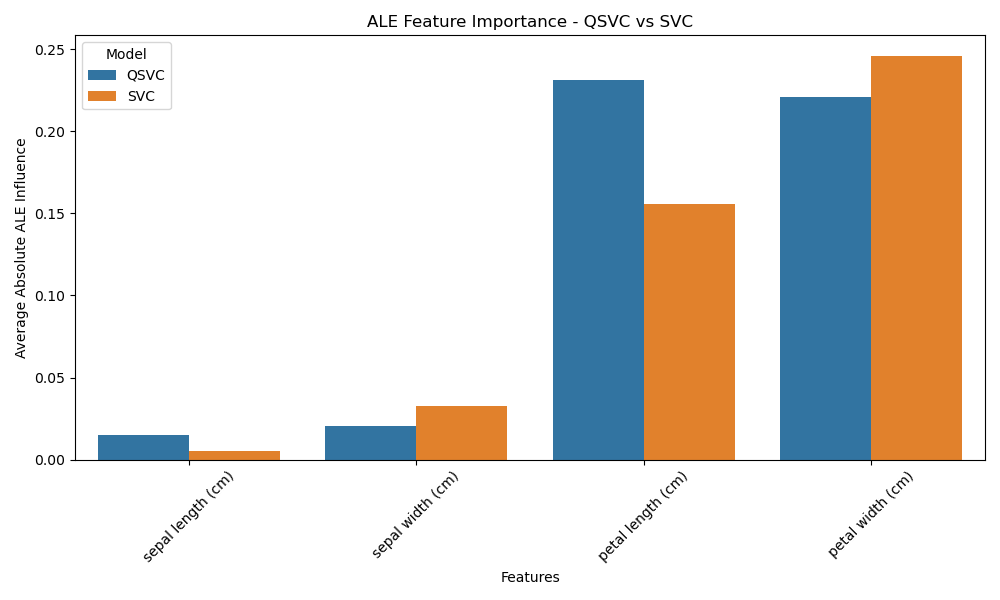} \caption[ALE Feature Importance]{ALE Importance.}
  \label{fig:ale_comparison}
\end{figure*}

\begin{figure}[h!]
\centering
  \includegraphics[width=1\linewidth]{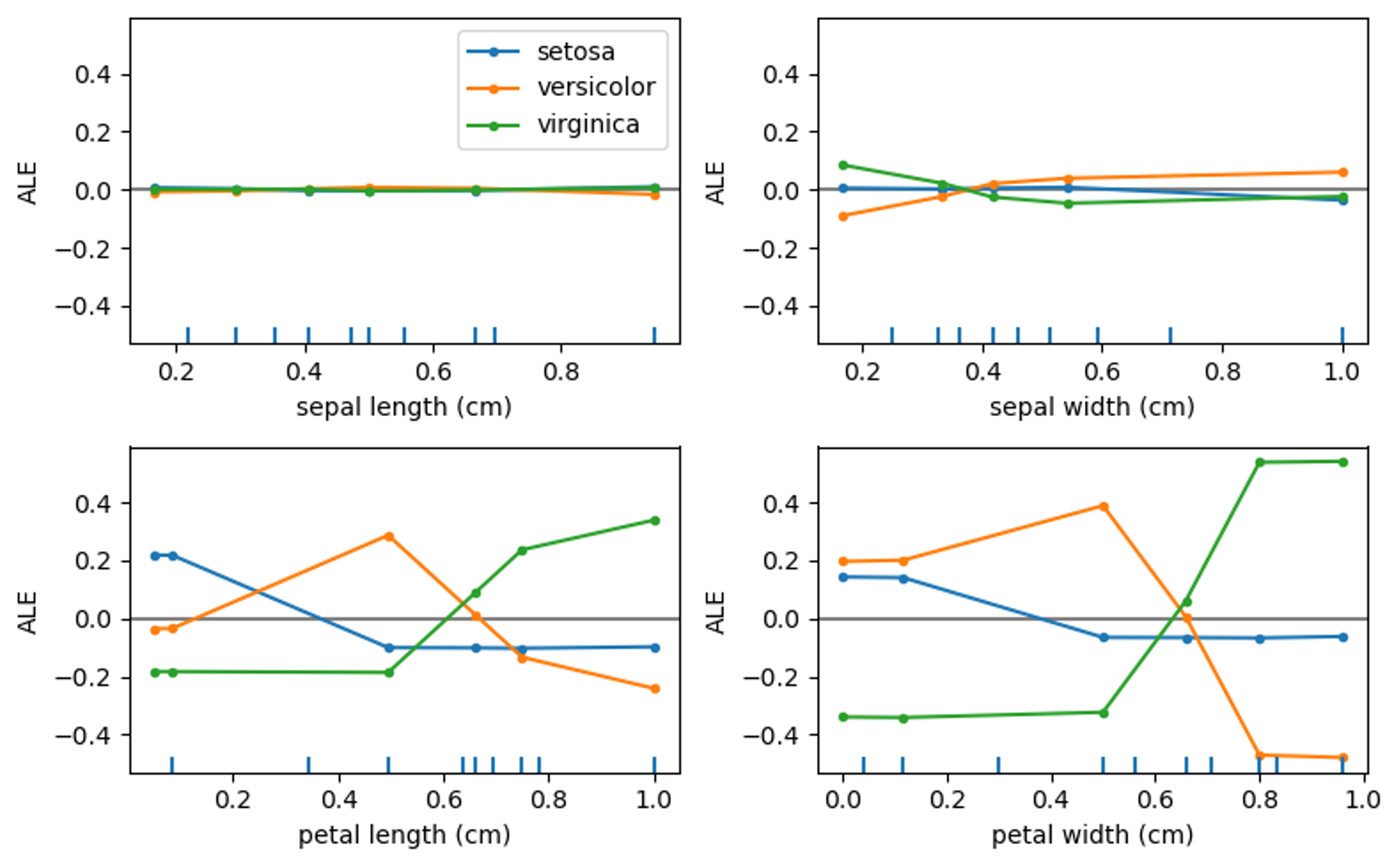} \caption[SVC ALE]{SVC ALE.}
  \label{fig:svc_ale_classes}
\end{figure}

\begin{figure}[h!]
\centering
  \includegraphics[width=1\linewidth]{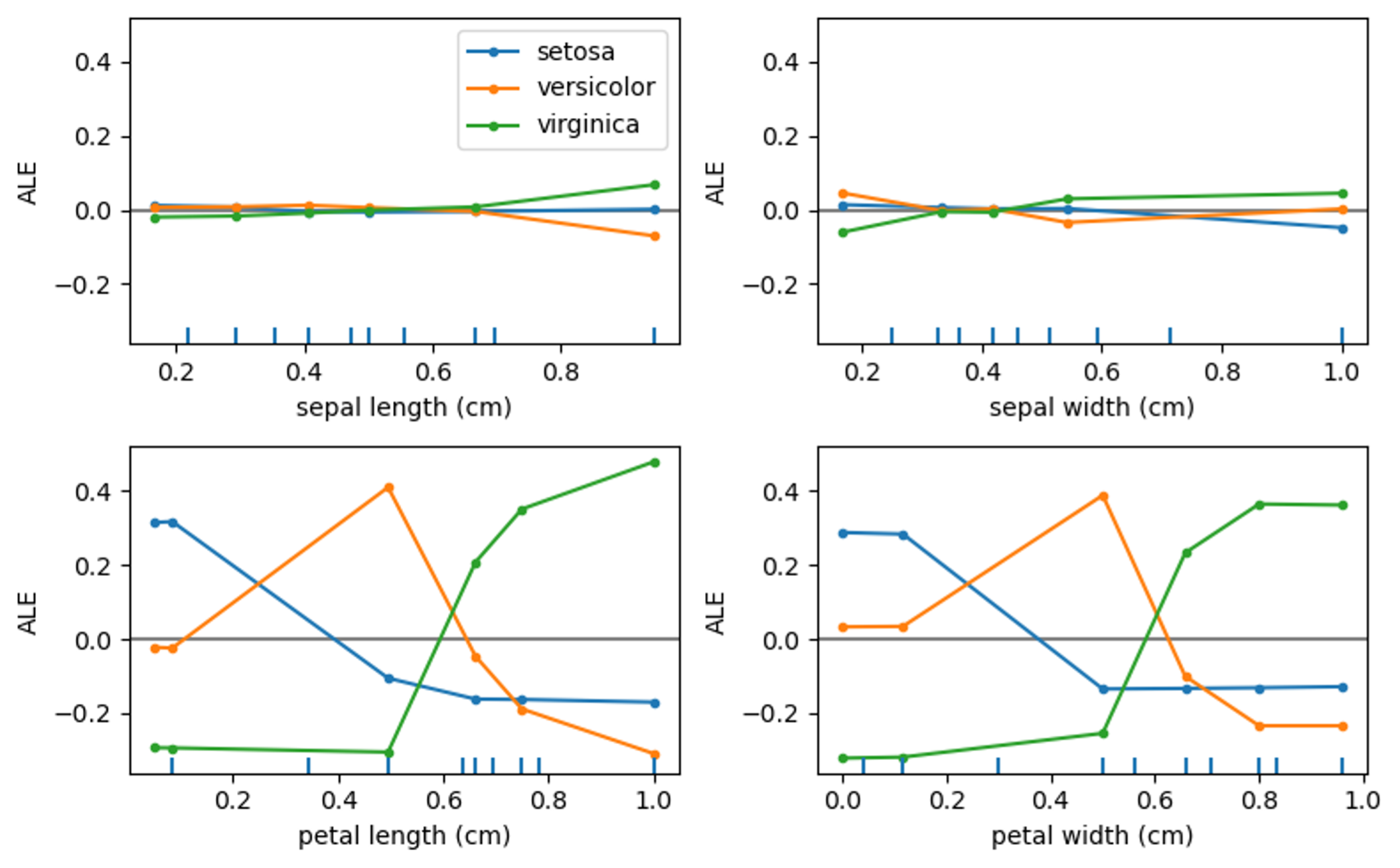} \caption[QSVC ALE]{QSVC ALE.}
  \label{fig:qsvc_ale_classes}
\end{figure}

 Plot \ref{fig:ale_comparison} shows that both models follow a similar distribution of importance, however the QSVC results with higher ALE influence in three out of the four features, with the largest increase in the importance of petal length. But overall the ALE values of both the SVC and QSVC are highly similar. \\
 A better way to utilise and understand ALE values is in plots \ref{fig:svc_ale_classes} and \ref{fig:qsvc_ale_classes}. These plots show the ALE values for SVC and QSVC respectively. We can interpret the values as the increase or decrease in the probability of predicting the class of interest. We can see that petal length and width are important features due to how much the ALE values change for each of the values. For example, we can see that the probability of predicting Virginica in both SVC and QSVC increases as petal length increases. We can also see in the QSVC that sepal width has more of an impact in the model prediction as values increase, while the same has very almost no effect in the SVC model.

\section{Explainability}
While feature importance can be useful for a general overview of what are the most important features in a model, more often than not we would like to see more specific explanations for a particular classification or individual result. The two methods used for this are Accumulated Local Effects and SHAP.

\subsubsection{ALE For Class Explanation}
We have shown in section \ref{section: ALE_FI} how ALE can be used to show how features affect a model's prediction, this can be inverted so that we can see how the different models differ in predicting a class. Given how we can see that the  Versicolor and Virginica classes overlap in the pairs plot \ref{fig:iris_pairs_plot}, we will investigate how ALE values can explain how the different models differ in predicting these classes. 

\subsubsection*{Versicolor Prediction Model Comparison}
\begin{figure}[h!]
  \includegraphics[width=1\linewidth]{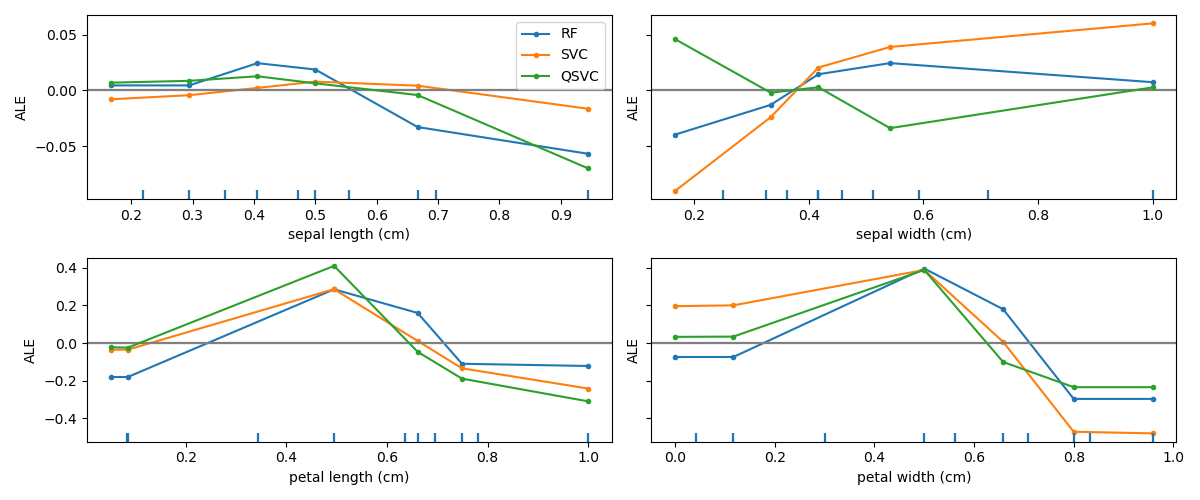} \caption[Versicolor ALE]{Versicolor ALE Values.}
  \label{fig:Versicolor_ale}
\end{figure}

In plot \ref{fig:Versicolor_ale} we can see how the ALE values for each feature differ by model when the class of interest is Versicolor. We can see that all three models follow similar trends for each of the features, however, we can see that a higher value of sepal width resulted in the QSVC increasing its probability of predicting the sample as Versicolor, while the same value had zero impact in the SVC and Random Forest models. A very interesting observation is also made concerning sepal length. While a high value of sepal length indicated a higher probability of being Versicolor, a higher value of sepal length decreased the probability of making that same classification. However, we can see the magnitude of these changes is only a fraction of the decision when compared to the ALE values for petal length and petal width, with the sepal features giving ALE values in the region of 0 to 0.05 compared to the petal features who's effects are in the range of 0 to 0.4, which strongly decrease the probability of predicting a  Versicolor class. \\
Plot \ref{fig:Versicolor_ale} show that the three models are most likely to predict that a case is Versicolor when the normalised values of petal length and width are approximately between 0.25cm and 0.65cm, with ALE scores decreasing at either side of this interval.

\subsubsection*{Virginica Prediction Model Comparison}

\begin{figure}[h!]
  \includegraphics[width=1\linewidth]{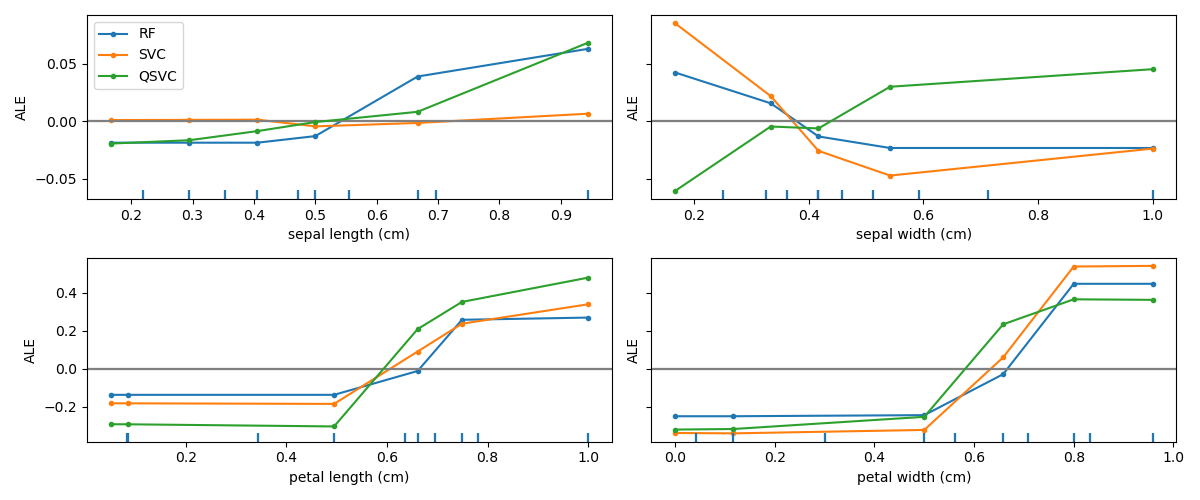} \caption[Virginica ALE]{Virginica ALE Values.}
  \label{fig:virginica_ale}
\end{figure}

How the ALE values fluctuate across the features and models is shown if plot \ref{fig:virginica_ale}. Similarly to the Versicolor analysis, we can see that the QSVC's ALE values increase with sepal length and width, but this time the effect is greatly magnified with the same pattern in petal length and width. What we could infer from this visualisation is that for high values of all four features, the QSVC model has a higher probability of predicting Virginica.\\

When the two plots \ref{fig:Versicolor_ale} and \ref{fig:virginica_ale} are viewed together we really see how petal width and length differentiate the model's predictions between the two classifications. With all three models significantly decreasing as petal measurements increase in predicting Versicolor and at the same time, we can see the ALE values for Virginica increasing. This could also help us further understand where the model may make mistakes, particularly when an instance of petal length and width is in approximately the 0.6 to 0.7 regions. \\

We can see in this analysis how much more information can be inferred from the data by ALE by both classical and quantum classification models when analysis goes beyond the generality of feature importance analysis. Such analysis could also be made using  SHAP values, which expands again on the ALE values by proving clear breakdowns of individual results. 

\subsubsection{SHAP For Case-Specific Predictions}
SHAP uses game theory coalition values to estimate the marginal contribution from a feature to a model's prediction. This is a method that can be used to quantify how features in a model contribute to individual prediction, rather than focusing on a feature importance method like Permutation Importance, which has its merits when we want to get an idea of how a model would typically work, however, these generalisations cannot quantify the effects in individual cases. 

The table \ref{tab:misclassification}, and the plot \ref{fig:test_pair_plot} show us that five out of the thirty test points are misclassified. We can see that point 4, which has a true label of Virginica was misclassified as Versicolor (label 1) by all models except the QSVC, this makes this an ideal candidate to use for further investigation. The feature values for point 4 can be found in table \ref{tab:4feature_table}.\\

\begin{table}[h]
\centering
\begin{tabular}{|l|c|}
\hline
\textbf{Feature} & \textbf{Value} \\
\hline
0 (Sepal Length) & 0.5 \\
\hline
1 (Sepal Width) & 0.25 \\
\hline
2 (Petal Length) & 0.78 \\
\hline
3 (Petal Width) & 0.54 \\
\hline
\end{tabular}
\caption{Features of the test point 4}
\label{tab:4feature_table}
\end{table}

\subsubsection*{SVC SHAP Values}
\begin{figure}[h!]
\centering
  \includegraphics[width=1\linewidth]{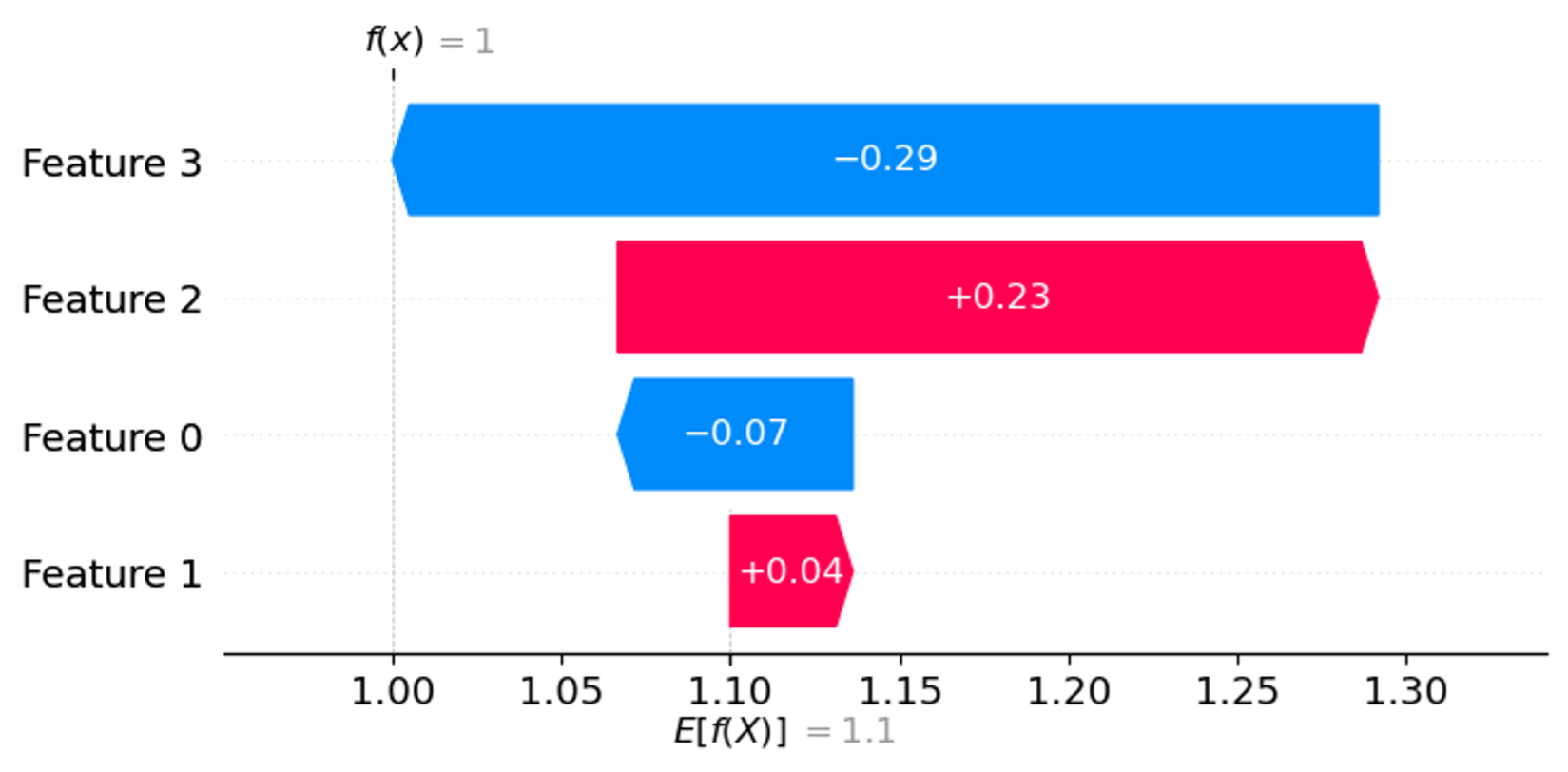} \caption[SVC SHAP]{SVC SHAP}
  \label{fig:svc_shap_waterfall}
\end{figure}

Figure \ref{fig:svc_shap_waterfall} is a waterfall plot of the SHAP values for sample 4. We can see that the expected output, or base value is 1.1, which would fall into the Versicolor class, we can see that Feature 3 (petal width) causes the largest contribution to this prediction, however the push is almost completely offset by feature 2 (petal width), this confirms what has been discussed previously, being petal width and length being the most important features in the SVC. The SHAP values quantify their importance in making this particular prediction.  We can see the effects of sepal length and width are minimal in comparison. The contrasting effects by features two and three that the model uses  could be explained that the model could be using the effect of petal width and length to cancel each other out in order to stay, on the base  classification and align with the classification of 1, Versicolor.\\

\subsubsection*{Random Forest SHAP Values}
\begin{figure}[h!]
\centering
  \includegraphics[width=1\linewidth]{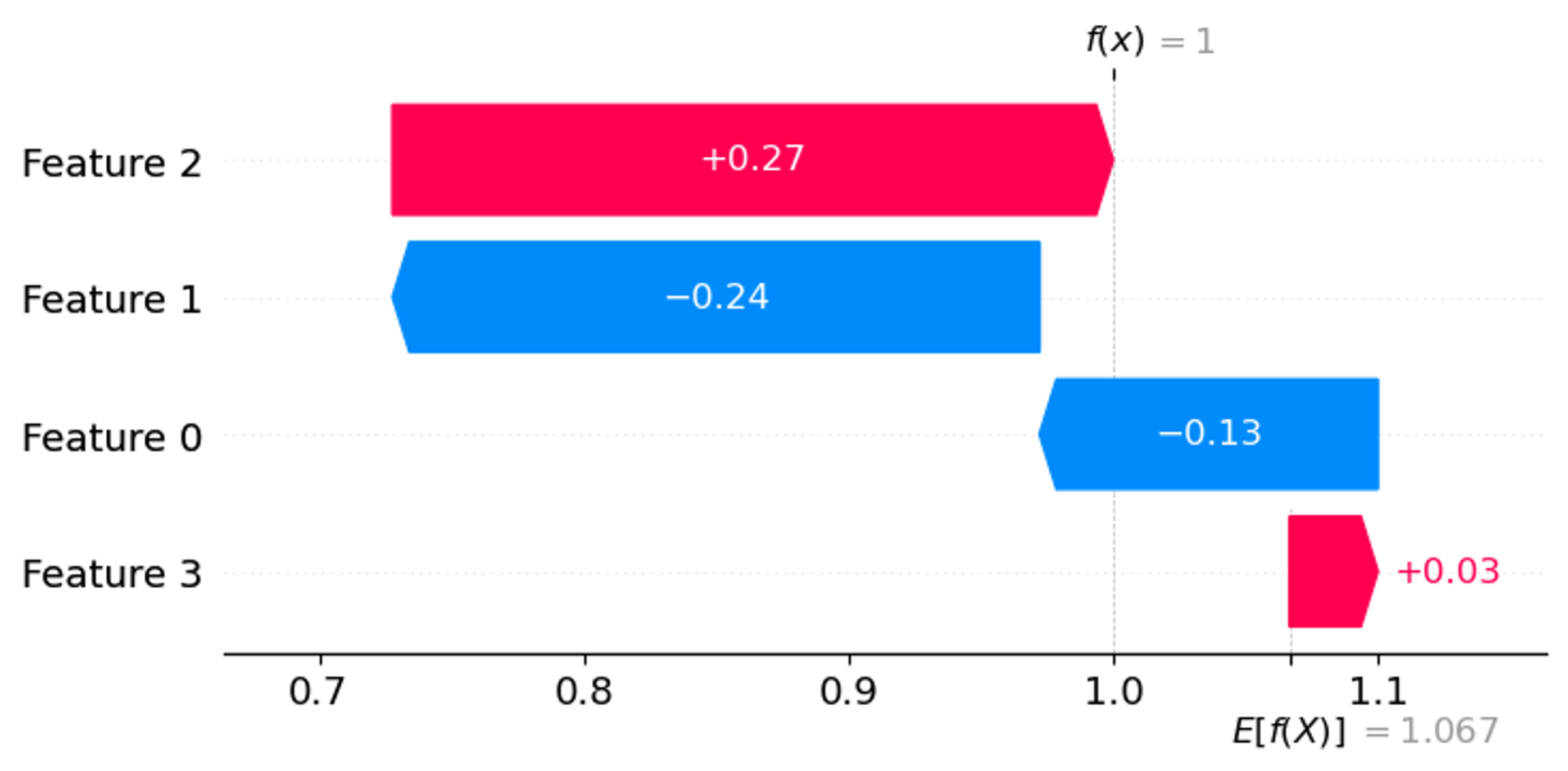} \caption[Random Forest Shap]{Random Forest SHAP}
  \label{fig:rf_shap_waterfall}
\end{figure}

\begin{figure}[h!]
\centering
  \includegraphics[width=1\linewidth]{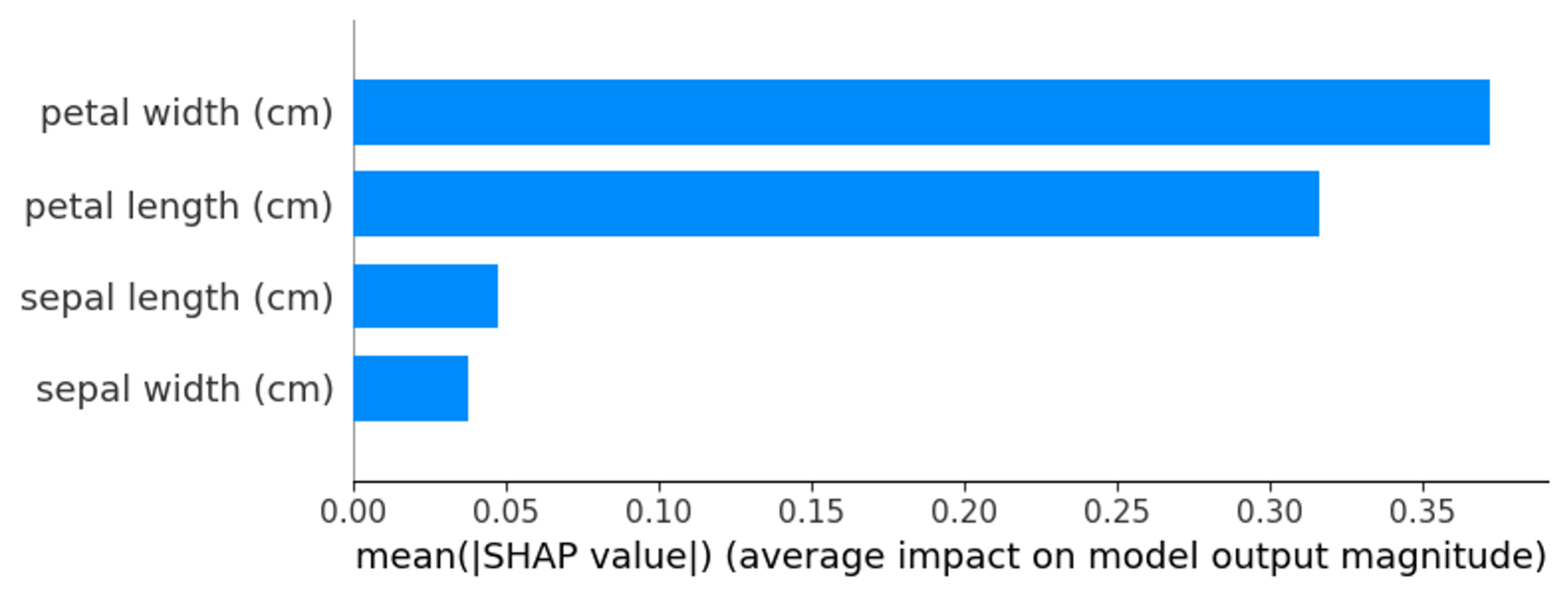} \caption[Random Forest Shap Feature Importance]{Random Forest SHAP Feature Importance}
  \label{fig:rf_shap_importance}
\end{figure}

ForWith respect to the Random Forest classifier, we can see that it tended to go towards a setosa classification, again we can see that petal length strongly pushes to the right, however in this model we can see that it is sepal width is a highly important feature for this case, which differs if we look at an aggregation of all the SHAP values in plot \ref{fig:rf_shap_importance},  where petal sepal width almost has no importance overall.\\ 

Incorrect classification aside, this shows how much SHAP values can demonstrate how important it is to be able to understand a models functionality on a case-by-case basis rather than using a general overview of a features importance.

\subsubsection*{QSVC SHAP Values}
\begin{figure}[h!]
\centering
  \includegraphics[width=1\linewidth]{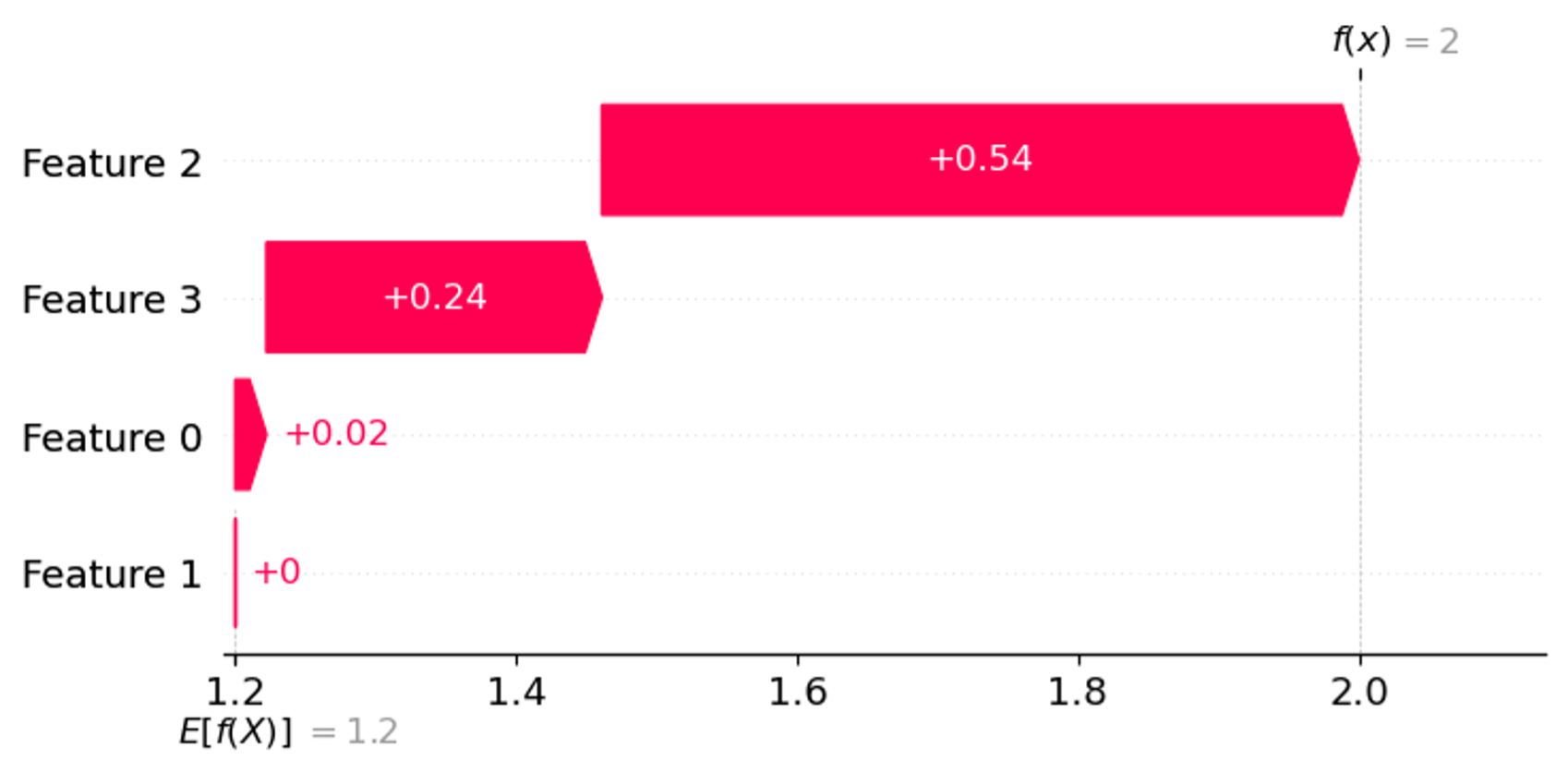} \caption[QSVC SHAP]{QSVC SHAP}
  \label{fig:qsvc_shap_waterfall}
\end{figure}

In our QSVC we can see that the model decided that all the feature values indicated a Virginica classification, with petal length and width being the two most important features by far, with sepal length only being assigned a SHAP value of 0.02 and sepal width not contributing at all in this classification. It is also an interesting thing to note that none of the feature values in the QSVC suggested that the prediction should change in the other direction.

\subsubsection{Summary of Analysis}
\subsubsection*{Feature Importance}
We have explored methods of feature importance across different models using three distinct methodologies: Leave One Out, Permutation Importance, and Accumulated Local Effects. Each method offered an insight into how features affect model performance and accuracy. We have then applied these methods to QML models.\\
LOO analysis highlighted the differential impact of features across models. For instance, SVC and Random Forest models showed an increase in accuracy upon the removal of petal features, suggesting a possible over-reliance or redundancy. In contrast, the VQC model exhibited significant performance drops, especially with the removal of petal width, indicating a high dependency on this feature. This disparity underscores the different ways models prioritize or handle feature dependencies.\\
Permutation importance further emphasized the significance of petal features in classical models, consistent with LOO results but provided additional details on feature interactions and their impact on model accuracy. The QSVC model's feature importance distribution appeared more aligned with classical models, however, we can observe some variations in sepal features' impact.\\
ALE uses probabilities to gauge a feature's impact on the model, however, due to constraints in the implementation of the VQC, this model had to be omitted from this method. The ALE plots demonstrated how the QSVC follows the trends seen in the classical models, but displays a higher sensitivity to the changes across all the features with almost no flat lines in any of the ALE plots \ref{fig:Versicolor_ale} \ref{fig:virginica_ale}.

\subsection*{Model Explainability}
Model explainability was addressed through ALE and SHAP analyses, focusing on how specific features influence individual predictions.
ALE analysis demonstrated the variable influence of features across different classes, providing insights into how models differentiate between the classes Versicolor and Virginica based on feature values. This analysis allows us to uncover a better understanding of how the decision-making processes in both classical and quantum models work.\\

SHAP values offered deeper insights into the contribution of individual features to specific predictions, highlighting the interplay of features in models like QSVC, which successfully identified the Virginica class in test point 4, which was misclassified by the other three models.

\section{Conclusion}
In this article we investigated different methods of feature importance (LOO, Permutation Importance, ALE) and explainability (ALE, SHAP) and applied these methods to quantum machine learning models, comparing the results to their classical counterpart. In this comparison, we have set a small milestone in this field of explainable quantum machine learning. This article displays that these methods can indeed be applied to QML models, and so are potential candidates as tools in the unexplored field of explainable quantum machine learning, or XQAI. It was also a highly interesting takeaway in that we haven't necessarily seen a reliable example of the quantum advantage in our QML models, probably due to the small scale of the data, but there are still applications to be explored through QML, for example, researching gradients and making quantum software ready for machine learning applications \cite{Schuld_2022}. \\
It is also important to remember that these are methods that were designed, tested, implemented and are used with almost exclusively classical machine learning tools in mind, meaning that there is likely a need for not only amending these tools but perhaps a need to come up with completely new quantum-specific methods of explainability. \cite{heese2023explaining}

The plans for this article are to take the QML models and train them with more complicated and larger, real-world data. This would further enhance the necessity of applying explainability techniques. To do this we would need to further optimise the quantum models, as well as design them optimally for running on a real quantum computer. The lack of models trained on an actual quantum computer was one of the greatest limitations of this article, as a simulator was used to avoid the extra time and complexity required, which was able to work due to the small scale of the data used. Due to the necessary time and resources this would take, attempting this on a larger dataset was beyond what was believed to be a realistic task for this article's time frame. It would also be required to explore more methods of quantum machine learning and algorithms such as Quantum Boltzmann Machines, and perhaps explore the possibility of quantum-specific explainability techniques.\\ \\
The notebooks containing the code, and the resources for this article can be found on the repository accessible at: 
https://github.com/LukePower01/ml-to-qml

\bibliographystyle{IEEEtran}
\bibliography{bare_jrnl_compsoc}




\end{document}